\documentclass[runningheads]{llncs}
\usepackage{graphicx}
\usepackage{amsmath}    
\usepackage{multirow}
\usepackage{amsmath}
\usepackage{amssymb}
\usepackage{appendix}
\usepackage{xargs} 
\usepackage{comment}
\usepackage{hyperref}
\usepackage{xcolor}
\usepackage{mdframed}
\definecolor{mycolor}{RGB}{173,216,230}
\definecolor{mytextcolor}{RGB}{0,0,205} % instruction background colors
\newmdenv[linewidth=0pt, fontcolor=mytextcolor]{prefixquote}

\usepackage[colorinlistoftodos,prependcaption,textsize=tiny]{todonotes}
\newcommandx{\hamed}[2][1=]{\todo[linecolor=blue,backgroundcolor=blue!50,bordercolor=black,#1]{Hamed: #2 }}
\newcommandx{\jennifer}[2][1=]{\todo[linecolor=red,backgroundcolor=red!60,bordercolor=black,#1]{Jennifer: #2 }}
\newcommandx{\soeren}[2][1=]{\todo[linecolor=green,backgroundcolor=green!80,bordercolor=black,#1]{Soeren: #2 }}
\usepackage{array}
\newcolumntype{P}[1]{>{\centering\arraybackslash}p{#1}}
\usepackage{rotating}

\newcolumntype{K}[2]{%
  >{\begin{turn}{#1}\begin{minipage}{#2}\small\raggedright\hspace{0pt}}l%
  <{\end{minipage}\end{turn}}%
}

\usepackage{graphicx}
\usepackage{url}

\begin{document}
\title{LLMs4OL: Large Language Models‌ for \\Ontology Learning}
\titlerunning{LLMs4OL: Large Language Models‌ for Ontology Learning}
\author{
Hamed Babaei Giglou\orcidID{0000-0003-3758-1454} \and Jennifer D’Souza\orcidID{0000-0002-6616-9509} \and S\"{o}ren Auer\orcidID{0000-0002-0698-2864}}
\authorrunning{Babaei Giglou et al.}
\institute{TIB Leibniz Information Centre for Science and Technology, Hannover, Germany\\
\email{\{hamed.babaei,jennifer.dsouza,auer\}@tib.eu}\\
% \institute{TIB Leibniz Information Centre for Science and Technology\\ Hannover, Germany\\
% \email{\{email1, email2, email3, ... \}@tib.eu}\\
}
\maketitle              % typeset the header of the contribution
\begin{abstract}

We propose the LLMs4OL approach, which utilizes Large Language Models (LLMs) for Ontology Learning (OL). LLMs have shown significant advancements in natural language processing, demonstrating their ability to capture complex language patterns in different knowledge domains. Our LLMs4OL paradigm investigates the following hypothesis: \textit{Can LLMs effectively apply their language pattern capturing capability to OL, which involves automatically extracting and structuring knowledge from natural language text?} To test this hypothesis, we conduct a comprehensive evaluation using the zero-shot prompting method. We evaluate nine different LLM model families for three main OL tasks: term typing, taxonomy discovery, and extraction of non-taxonomic relations. Additionally, the evaluations encompass diverse genres of ontological knowledge, including lexicosemantic knowledge in WordNet, geographical knowledge in GeoNames, and medical knowledge in UMLS.

%We introduce the LLMs4OL approach for leveraging Large Language Models (LLMs) for Ontology Learning (OL). LLMs have made strides in natural language processing, having shown great potential in capturing complex patterns in language data across various domains of knowledge. In this context, the LLMs4OL paradigm tests the following hypothesis for the first time: \textit{Does the capability of LLMs to capture intricate linguistic relationships translate effectively to OL, given that OL mainly relies on automatically extracting and structuring knowledge from natural language text?} As such, a comprehensive set of evaluations adopting the zero-shot prompting method across five different flavors of LLMs are presented for the following three main facets of OL: term typing, taxonomy discovery, and non-taxonomic relations extraction. Furthermore, the evaluations cover various genres of ontological knowledge -- lexicosemantic knowledge in WordNet, geographical knowledge in GeoNames, and medical knowledge in UMLS.

The obtained empirical results show that foundational LLMs are not sufficiently suitable for ontology construction that entails a high degree of reasoning skills and domain expertise. Nevertheless, when effectively fine-tuned they just might work as suitable assistants, alleviating the knowledge acquisition bottleneck, for ontology construction.

% The abstract should briefly summarize the contents of the paper in
% 15--250 words.

\end{abstract}
\section{Introduction}
% \hamed[inline]{some introductions + backgrounds, intro for the research question, research questions, what we did for answering the research question, finally paper ognization}
%artificial general intelligence (AGI)
%Knowledge Acquisition is still the bottleneck in building many kinds of applications. This problem is especially acute for the task of Ontology Learning (OL) which is traditionally done by humans therefore costly. Furthermore, OL is also time-consuming when done by humans only causing the KA bottleneck problem to become more acute.
%there is a need for an automatic acquisition methodology.
%Ontology learning (OL) \cite{ontologylearning} is the process of automatically extracting and organizing knowledge from unstructured or semi-structured text into a structured representation, such as an ontology. LLMs can assist in this process by generating high-quality entities and coherent descriptions of entities and their relationships, which can then be used to populate an ontology. 
%Ontology learning from text is the process of identifying terms, concepts, relations, and optionally, axioms from textual information and using them to construct an ontology~\cite{konys2019knowledge}.
Ontology Learning (OL) is an important field of research in artificial intelligence (AI) and knowledge engineering, as it addresses the challenge of knowledge acquisition and representation in a variety of domains. OL involves automatically identifying terms, types, relations, and potentially axioms from textual information to construct an ontology~\cite{konys2019knowledge}. Numerous examples of human-expert created ontologies exist, ranging from general-purpose ontologies to domain-specific ones, e.g., Unified Medical Language System (UMLS)~\cite{bodenreider2004unified}, WordNet~\cite{miller1995wordnet}, GeoNames~\cite{rebele2016yago}, Dublin Core Metadata Initiative (DCMI)~\cite{weibel2000dublin}, schema.org~\cite{guha2016schema}, etc. Traditional ontology creation relies on manual specification by domain experts, which can be time-consuming, costly, error-prone, and impractical when knowledge constantly evolves or domain experts are unavailable. Consequently, OL techniques have emerged to automatically acquire knowledge from unstructured or semi-structured sources, such as text documents and the web, and transform it into a structured ontology. A quick review of the field shows that traditional approaches to OL are based on lexico-syntactic pattern mining and clustering~\cite{xu2002domain,missikoff2002usable,lonsdale2002peppering,khan2002ontology,alfonseca2002unsupervised,hahn2001joint,wagner2000enriching,roux2000ontology,kietz2000method,agirre2000enriching,hwang1999incompletely,hearst1998automated}. In contrast, recent advances in natural language processing (NLP) through Large Language Models (LLMs)~\cite{chatgpt} offer a promising alternative to traditional OL methods. The ultimate goal of OL is to provide a cost-effective and scalable solution for knowledge acquisition and representation, enabling more efficient and effective decision-making in a range of domains. To this end, we introduce the LLMs4OL paradigm and empirically ground it as a foundational first step.

Currently, there is no research explicitly training LLMs for OL. Thus to test LLMs for OL for the first time, we made some experimental considerations. The first being: \textit{Do the characteristics of LLMs justify ontology learning?} First, LLMs are trained on extensive and diverse text, similar to domain-specific knowledge bases~\cite{petroni2019language}. This aligns with the need for ontology developers to have extensive domain knowledge. Second, LLMs are built on the core technology of transformers that have enabled their higher language modeling complexity by facilitating the rapid scaling of their parameters. These parameters represent connections between words, enabling LLMs to comprehend the meaning of unstructured text like sentences or paragraphs. Further, by extrapolating complex linguistic patterns from word connections, LLMs exhibit human-like response capabilities across various tasks, as observed in the field of ``emergent'' AI. This behavior entails performing tasks beyond their explicit training, such as generating executable code, diverse genre text, and accurate text summaries~\cite{srivastava2022beyond,wei2022emergent}. Such ability of LLMs to extrapolate patterns from simple word connections, encoding language semantics, is crucial for OL. Ontologies often rely on analyzing and extrapolating structured information connections, such as term-type taxonomies and relations, from unstructured text \cite{gruber1995toward}.
%Such ability shown by LLMs to extrapolate patterns from basic word connections encoding the semantics of language is critical for OL, as ontologies are often built based on analyzing and extrapolating structured information connections such as term-type taxonomies and relations from unstructured text~\cite{gruber1995toward}. %Consequently, LLMs excel at crystallizing knowledge and patterns from extensive text sources. 
%As OL aims to extract a shared conceptualization by understanding types and relationships from diverse sources~\cite{gruber1995toward}, LLMs fulfill this requirement. 
%Thus the LLMs4OL hypothesis of the fruitful application of LLMs for OL seemed conceptually justified.
Thus LLMs4OL hypothesis of LLMs' fruitful application for OL appeared conceptually justified.

LLMs are being developed at a rapid pace. At the time of writing of this work, at least 60 different LLMs are reported~\cite{amatriain2023transformer}. This led to our second main experimental consideration. \textit{Which LLMs to test for the LLMs4OL task hypothesis?} Empirical validation of various LLMs is crucial for NLP advancements and selecting suitable models for research tasks. Despite impressive performances in diverse NLP tasks, LLM effectiveness varies. For the foundational groundwork of LLMs4OL, we comprehensively selected eight diverse model families based on architecture and reported state-of-the-art performances at the time of this writing. The three main LLM architectures are encoder, decoder, and encoder-decoder. The selected LLMs for validation are: BERT~\cite{bert} (encoder-only); BLOOM~\cite{bloom}, MetaAI's LLaMA~\cite{llama}, OpenAI's GPT-3~\cite{gpt3}, GPT-3.5~\cite{chatgpt}, GPT-4~\cite{gpt4} (all decoder-only); and BART~\cite{bart} and Google's Flan-T5~\cite{flant5} (encoder-decoder).
Recent studies show that BERT excels in text classification and named entity recognition~\cite{bert}, BART is effective in text generation and summarization~\cite{bart}, and LLaMA demonstrates high accuracy in various NLP tasks, including reasoning, question answering, and code generation~\cite{llama}. Flan-T5 emphasizes instruction tuning and exhibits strong multi-task performance~\cite{flant5}. BLOOM's unique multilingual approach achieves robust performance in tasks like text classification and sequence tagging~\cite{bloom}. Lastly, the GPT series stands out for its human-like text generation abilities~\cite{gpt3,chatgpt,gpt4}. In this work, we aim to comprehensively unify these LLMs for their effectiveness under the LLMs4OL paradigm for the first time. 

%what are we trying to achieve or show the community with this work?
With the two experimental considerations in place, we now introduce the LLMs4OL paradigm and highlight our contributions. LLMs4OL is centered around the development of ontologies that comprise the following primitives~\cite{maedche2001ontology}: \textbf{1.} a set of strings that describe terminological lexical entries $L$ for conceptual types; \textbf{2.} a set of conceptual types $T$; \textbf{3.} a taxonomy of types in a hierarchy $H_{T}$; \textbf{4.} a set of non-taxonomic relations $R$ described by their domain and range restrictions arranged in a heterarchy of relations $H_{R}$; and \textbf{5.} a set of axioms $A$ that describe additional constraints on the ontology and make implicit facts explicit. The LLMs4OL paradigm, introduced in this work, addresses three core aspects of OL as tasks, outlined as the following research questions (RQs). %In this context, the LLMs4OL paradigm laid out for the first time in this work selectively tackles three core aspects of OL as tasks. They are presented as the following research questions (RQs).

\begin{itemize}
    \item \textbf{RQ1}: \textit{Term Typing Task} -- How effective are LLMs for automated type discovery to construct an ontology?
    \item \textbf{RQ2}: \textit{Type Taxonomy Discovery Task} -- How effective are LLMs to recognize a type taxonomy i.e. the ``is-a'' hierarchy between types?
    \item \textbf{RQ3}: \textit{Type Non-Taxonomic Relation Extraction Task} -- How effective are LLMs to discover non-taxonomic relations between types?
\end{itemize}

The diversity of the empirical tests of this work are not only w.r.t. LLMs considered, but also the ontological knowledge domains tested for. %Since LLMs are constructed over diverse domains or genres of text, an assumption made in the LLMs4OL paradigm is that they are just as simultaneously capable of OL over diverse knowledge domains. 
Specifically, we test LLMs for lexico-semantic knowledge in WordNet~\cite{miller1995wordnet}, geographical knowledge in GeoNames~\cite{geonames}, biomedical knowledge in UMLS~\cite{umls}, and web content type representations in schema.org~\cite{schemaorg}. For our empirical validation of LLMs4OL, we seize the opportunity to include PubMedBERT~\cite{pubmedbert}, a domain-specific LLM designed solely for the biomedical domain and thus applicable only to UMLS. This addition complements the eight domain-independent model families introduced earlier as a ninth model type. Summarily, our main contributions are:

\begin{itemize}
    \item The LLMs4OL task paradigm as a conceptual framework for leveraging LLMs for OL.
    \item An implementation of the LLMs4OL concept leveraging tailored prompt templates for zero-shot OL in the context of three specific tasks, viz. term typing, type taxonomic relation discovery, and type non-taxonomic relation discovery. These tasks are evaluated across unique ontological sources well-known in the community. Our code source with templates and datasets per task are released here \url{https://github.com/HamedBabaei/LLMs4OL}.
    \item A thorough out-of-the-box empirical evaluation of eight state-of-the-art domain-independent LLM types (10 models) and a ninth biomedical domain-specific LLM type (11th model) for their suitability to the various OL tasks considered in this work. Furthermore, the most effective overall LLM is finetuned and subsequently finetuned LLM results are reported for our three OL tasks.
\end{itemize}

\section{Related Work} 
\label{rel-work}

There are three avenues of related research: ontology learning from text, prompting LLMs for knowledge, and LLM prompting methods or prompt engineering.

\noindent{\textbf{Ontology Learning from Text.}} One of the earliest approaches~\cite{hearst1998automated} used lexicosyntactic patterns to extract new lexicosemantic concepts and relations from large collections of unstructured text, enhancing WordNet~\cite{miller1995wordnet}. 
WordNet is a lexical database comprising a lexical ontology of concepts (nouns, verbs, etc.) and lexico-semantic relations (synonymy, hyponymy, etc.). Hwang~\cite{hwang1999incompletely} proposed an alternative approach for constructing a dynamic ontology specific to an application domain. The method involved iteratively discovering types and taxonomy from unstructured text using a seed set of terms representing high-level domain types. In each iteration, newly discovered specialized types were incorporated, and the algorithm detected relations between linguistic features. The approach utilized a simple ontology algebra based on inheritance hierarchy and set operations. Agirre et al.\cite{agirre2000enriching} enhanced WordNet by extracting topically related words from web documents. This unique approach added topical signatures to enrich WordNet. Kietz et al.\cite{kietz2000method} introduced the On-To-Knowledge system, which utilized a generic core ontology like GermaNet~\cite{hamp1997germanet} or WordNet as the foundational structure. It aimed to discover a domain-specific ontology from corporate intranet text resources. For concept extraction and pruning, it employed statistical term frequency count heuristics, while association rules were applied for relation identification in corporate texts. Roux et al.\cite{roux2000ontology} proposed a method to expand a genetics ontology by reusing existing domain ontologies and enhancing concepts through verb patterns extracted from unstructured text. Their system utilized linguistic tools like part-of-speech taggers and syntactic parsers. Wagner \cite{wagner2000enriching} employed statistical analysis of corpora to enrich WordNet in non-English languages by discovering relations, adding new terms to concepts, and acquiring concepts through the automatic acquisition of verb preferences. Moldovan and Girju~\cite{moldovan2001interactive} introduced the Knowledge Acquisition from Text (KAT) system to enrich WordNet's finance domain coverage. Their method involved four stages: (1) discovering new concepts from a seed set of terms, expanding the concept list using dictionaries; (2) identifying lexical patterns from new concepts; (3) discovering relations from lexical patterns; and (4) integrating extracted information into WordNet using a knowledge classification algorithm. In \cite{alfonseca2002unsupervised}, an unsupervised method is presented to enhance ontologies with domain-specific information using NLP techniques such as NER and WSD. The method utilizes a general NER system to uncover a taxonomic hierarchy and employs WSD to enrich existing synsets by querying the internet for new terms and disambiguating them through cooccurrence frequency. Khan and Luo~\cite{khan2002ontology} employed clustering techniques to find new terms, utilizing WordNet for typing. They used the self-organizing tree algorithm~\cite{dopazo1997phylogenetic}, inspired by molecular evolution, to establish an ontology hierarchy. Additionally, Xu et al.~\cite{xu2002domain} focused on automatically acquiring domain-specific terms and relations through a TFIDF-based single-word term classifier, a lexico-syntactic pattern finder based on known relations and collocations, and a relation extractor utilizing discovered lexico-syntactic patterns.

Predominantly, the approaches for OL~\cite{wkatrobski2020ontology} that stand out so far are based on lexico-syntactic patterns for term and relation extraction as well as clustering for type discovery. Otherwise, they build on seed-term-based bootstrapping methods. The reader is referred to further detailed reviews~\cite{r3-soa1,r3-soa2} on this theme for a comprehensive overall methodological picture for OL. Traditional NLP was defined by modular pipelines by which machines were equipped step-wise with annotations at the linguistic, syntactic, and semantic levels to process text. LLMs have ushered in a new era of possibilities for AI systems that obviate the need for modular NLP systems to understand natural language which we tap into for the first time for the OL task in this work.

\noindent{\textbf{Prompting LLMs for Knowledge.}} LLMs can process and retrieve facts based on their knowledge which makes them good zero-shot learners for various NLP tasks. Prompting LLMs means feeding an input $x$ using a \textit{template function} $f_{prompt}(x)$, a textual string prompt input that has some unfilled slots, and then the LLMs are used to probabilistically fill the unfilled information to obtain a final string $x^{\prime}$, from which the final output $y$ can be derived~\cite{prompting}. The LAMA: LAnguage Model Analysis~\cite{lms-as-kb} benchmark has been introduced as a probing technique for analyzing the factual and commonsense knowledge contained in unidirectional LMs (i.e. Transformer-XL~\cite{dai2019transformerxl}) and bidirectional LMs (i.e. BERT and ELMo~\cite{peters-etal-2018-deep}) with cloze prompt templates from knowledge triples. They demonstrated the potential of pre-trained language models (PLMs) in probing facts -- where facts are taken into account as subject-relation-object triples or question-answer pairs -- with querying LLMs by converting facts into a cloze template which is used as an input for the LM to fill the missing token. Further studies extended LAMA by the automated discovery of prompts~\cite{jiang-etal-2020-know}, finetuning LLMs for better probing~\cite{akkalyoncu-yilmaz-etal-2019-applying,levy-etal-2017-zero,yang2019simple}, or a purely unsupervised way of probing knowledge from LMs~\cite{petroni2020how}. These studies analyzed LLMs for their ability to encode various linguistic and non-linguistic facts. This analysis was limited to predefined facts that reinforce the traditional linguistic knowledge of the LLMs, and as a result do not reflect how concepts are learned by the LLMs. In response to this limitation, Dalvi et al.~\cite{dalvi2022discovering} put forward a proposal to explore and examine the latent concepts learned by LLMs, offering a fresh perspective on BERT. They defined concepts as ``a group of words that are meaningful,'' i.e. that can be clustered based on relations such as lexical, morphological, etc. In another study~\cite{sajjad2022analyzing}, they propose the framework \textit{ConceptX} by extending their studies on seven LLMs in latent space analysis with the alignment of the grouped concepts to human-defined concepts. These works show that using LLMs and accessing the concept's latent spaces, allows us to group concepts and align them to predefined types and type relations discovery.   

\noindent{\textbf{Prompt Engineering.}} As a novel discipline, prompt engineering focuses on designing optimal instructions for LLMs to enable successful task performance. Standard prompting \cite{wei2022finetuned} represents a fundamental approach for instructing LLMs. It allows users to craft their own customized ``self-designed prompts'' to effectively interact with LLMs \cite{gpt3} and prompt them to respond to the given prompt instruction straightaway with an answer. Consider the manually crafted FLAN collection~\cite{longpre2023flan} addressing diverse NLP tasks other than OL as an exemplar. Notably, the nature of some problems naturally encompass a step-by-step thought process for arriving at the answer. In other words, the problem to be solved can be decomposed as a series of preceding intermediate steps before arriving at the final solution. E.g., arithmetic or reasoning problems. Toward explainability and providing language models in a sense ``time to think'' helping it respond more accurately, there are advanced prompt engineering methods as well. As a first, as per the Chain-of-Thought (CoT) \cite{NEURIPS20229d560961} prompting method, the prompt instruction is so crafted that the LLM is instructed to break down complex tasks as a series of incremental steps leading to the solution. This helps the LLM to reason step-by-step and arrive at a more accurate and logical conclusion. On the other hand Tree-of-Thoughts (ToT) \cite{yao2023tree} has been introduced for tasks that require exploration or strategic lookahead. ToT generalizes over CoT prompting by exploring thoughts that serve as intermediate steps for general problem-solving with LLMs. Both CoT and ToT unlock complex reasoning capabilities through intermediate reasoning steps in combination with few-shot or zero-shot \cite{kojima2023large} prompting. Another approach for solving more complex tasks is using decomposed prompting \cite{khot2023decomposed}, where we can further decompose tasks that are hard for LLMs into simpler solvable sub-tasks and delegate these to sub-task-specific LLMs.

%However, tackling more complex tasks requires a deeper level of knowledge than what standard prompting permits. Particularly, reasoning tasks in NLP involve models performing logical inference and drawing conclusions from textual data.

Given the LLMs4OL task paradigm introduced in this work, complex prompting is not a primary concern, as our current focus is on the initial exploration of the task to identify the areas where we need further improvement. We want to understand how much we have accomplished so far before delving into more complex techniques like CoT, ToT, and decomposed prompting. Once we have a clearer picture of the model's capabilities and limitations in a standard prompting setting, we can then consider other than standard prompt engineering approaches by formulating OL as a stepwise reasoning task.

\section{The LLMs4OL Task Paradigm}
\begin{figure}[tb]
\includegraphics[width=\textwidth]{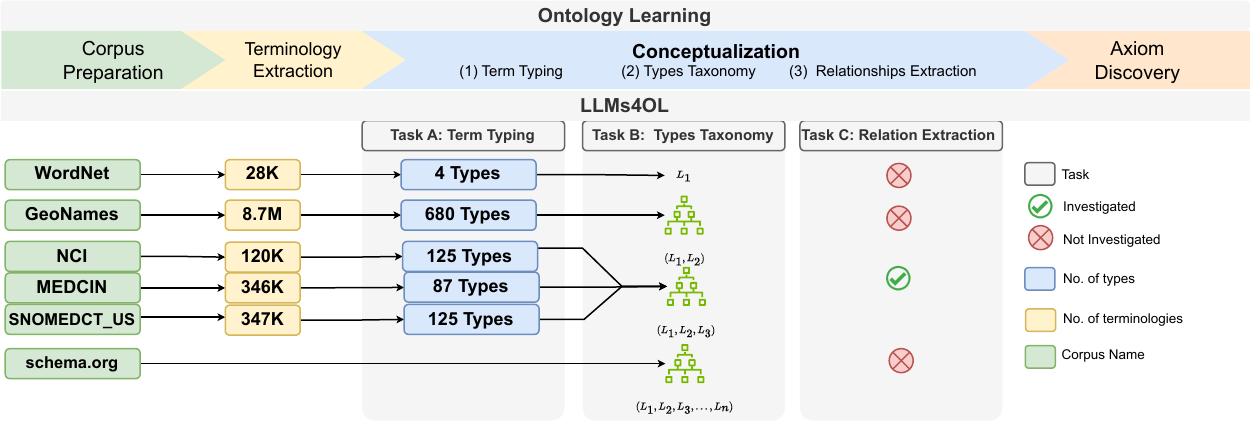}
%\vspace{-0.7cm}
\caption{The LLMs4OL task paradigm is an end-to-end framework for ontology learning in various knowledge domains, i.e. lexicosemantics (WordNet), geography (GeoNames), biomedicine (NCI, MEDICIN, SNOMEDCT), and web content types (schema.org). The three OL tasks empirically validated in this work are depicted within the blue arrow, aligned with the greater LLMs4OL paradigm.
%The LLMs4OL (Large Language Models for Ontology Learning) task paradigm is an end-to-end conceptual framework for learning ontologies in different knowledge domains, i.e. lexicosemantics -- WN18RR (WordNet), geography -- GeoNames, biomedicine -- NCI, MEDICIN, SNOMEDCT, and web content types -- Schema.Org. The tasks within the blue arrow are the three OL tasks empirically validated in this work in the context of the greater LLMs4OL paradigm.
} \label{llms4ol}
%\vspace{-5mm}
%\vspace{-0.7cm}
\end{figure}

The Large Language Models for Ontology Learning (LLMs4OL) task paradigm offers a conceptual framework to accelerate the time-consuming and expensive construction of ontologies exclusively by domain experts to a level playing field involving powerful AI methods such as LLMs for high-quality OL results; consequently and ideally involving domains experts only in validation cycles. In theory, with the right formulations, all tasks pertinent to OL fit within the LLMs4OL task paradigm. OL tasks are based on ontology primitives~\cite{maedche2001ontology}, including lexical entries $L$, conceptual types $T$, a hierarchical taxonomy of types $H_T$, non-taxonomic relations $R$ in a heterarchy $H_R$, and a set of axioms $A$ to describe the ontology's constraints and inference rules.
To address these primitives, OL tasks~\cite{noy2001ontology} include: 1) Corpus preparation - selecting and collecting source texts for ontology building. 2) Terminology extraction - identifying and extracting relevant terms. 3) Term typing - grouping similar terms into conceptual types. 4) Taxonomy construction - establishing "is-a" hierarchies between types. 5) Relationship extraction - identifying semantic relationships beyond "is-a." 6) Axiom discovery - finding constraints and inference rules for the ontology. This set of six tasks forms the LLMs4OL task paradigm. See \autoref{llms4ol} for the proposed LLMs4OL conceptual framework.

In this work, we empirically ground three core OL tasks using LLMs as a foundational basis for future research. However, traditional AI paradigms rely on testing models only on explicitly trained tasks, which is not the case for LLMs. Instead, we test LLMs for OL as an "\href{https://www.quantamagazine.org/the-unpredictable-abilities-emerging-from-large-ai-models-20230316/}{emergent}" behavior~\cite{srivastava2022beyond,wei2022emergent}, where they demonstrate the capacity to generate responses on a wide range of tasks despite lacking explicit training. The key to unraveling the emergent abilities of LLMs is to prompt them for their knowledge, as popularized by GPT-3~\cite{gpt3}, via carefully designed prompts. As discussed earlier (see \autoref{rel-work}), prompt engineering for LLMs is a new AI sub-discipline. In this process, a pre-trained language model receives a prompt, such as a natural language statement, to generate responses without further training or gradient updates to its parameters~\cite{prompting}. Prompts can be designed in two main types based on the underlying LLM pretraining objective: cloze prompts~\cite{petroni2019language,cui2021template}, which involve filling in blanks in an incomplete sentence or passage and suit masked language modeling pre-training; and prefix prompts~\cite{li2021prefix,lester2021power}, which generate text following a given starting phrase and offer more design adaptability to the underlying model. The earlier introduced LLMs4OL paradigm is empirically validated for three select OL tasks using respective prompt functions $f_{prompt}(.)$ suited to each task and model. %Thus, the earlier introduced LLMs4OL paradigm is empirically validated for the three select tasks pertaining to OL based on respective prompt functions $f_{prompt}(.)$ suited to the task and model at hand.

%As discussed earlier in \autoref{rel-work}, prompt engineering for LLMs is now a new AI sub-discipline. In prompting, a pre-trained language model is given a prompt (e.g. a natural language statement or instruction) of a task and completes the response without any further training or gradient updates to its parameters~\cite{prompting}. Based on the underlying LLM pretraining objective, prompts can be designed as one of two main types: cloze prompts~\cite{petroni2019language,cui2021template}, or prefix prompts~\cite{li2021prefix,lester2021power}. Cloze prompts involve generating text by filling in the blanks in a given incomplete sentence or passage and are best suited to models pre-trained with the masked language modeling objective; while prefix prompts involve generating text that follows a given starting phrase or sentence and are otherwise more adaptable in design to the underlying model.

%formalize the definitions of prompt-based zero-shot testing of LLMs
%Many of these emergent behaviors illustrate “zero-shot” or “few-shot” learning, which describes an LLM’s ability to solve problems it has never — or rarely — seen before.

%the effectiveness of probing language models is based on prompt design

\noindent{\textbf{Task A -} \underline{\textit{Term Typing}}.} A generalized type is discovered for a lexical term. 

The generic cloze prompt template is $f_{c-prompt}^{A}(L) := [S?].\;[L]\;[P_{domain}]\;is\;a\;\\\;[MASK].$ where $S$ is an optional context sentence, $L$ is the lexical term prompted for, $P_{domain}$ is a domain specification, and the special $MASK$ token is the type output expected from the model. Since prompt design is an important factor that determines how the LLM responds, eight different prompt template instantiations of the generic template were leveraged with final results reported for the best template. E.g., if WordNet is the base ontology, the part-of-speech type for the lexical term is prompted. In this case, template 1 is ``[S]. [L] POS is a [MASK].'' Note here ``$[P_{domain}]$'' is POS. Template 2 is ``[S]. [L] part of speech is a [MASK].'' Note here ``$[P_{domain}]$'' is ``part of speech.'' In a similar manner, eight different prompt variants from the generic template were created. However, the specification of ``$[P_{domain}]$'' depended on the ontology's knowledge domain.

%the specification of ``$[P_{domain}]$'' was contingent on the knowledge domain of the selected ontology.

The prefix prompt template reuses the cloze prompt template but appends an additional ``instruction'' sentence and replaces the special [MASK] token with a blank or a ``?'' symbol. Generically, it is $f_{p-prompt}^{A}(T) = [instruction] + f_{c-prompt}^{A}(T)$, where the instruction is ``Perform a sentence completion on the following sentence:'' Based on the eight variations created from the generic cloze template prompt, subsequently eight template variations were created for the prefix prompting of the LLMs as well with best template results reported.

\noindent{\textbf{Task B -} \underline{\textit{Taxonomy Discovery}}.} Here a taxonomic hierarchy between pairs of types is discovered.

The generic cloze prompt template is $f_{c-prompt}^{B}(a, b) := [a|b]\;is\;[P_{hierarchy}]\;of\;\\\;[b|a].\; This\;statement\;is\;[MASK]. $ Where $(a,b)$ or $(b,a)$ are type pairs, $P_{hierarchy}$ indicates superclass relations if the template is initialized for top-down taxonomy discovery, otherwise indicates subclass relations if the template is initialized for bottom-up taxonomy discovery. In Task B, the expected model output for the special [MASK] token for a given type pair was true or false. 

Similar to term typing, eight template variations of the generic template were created. Four of which were predicated on the top-down taxonomy discovery. E.g., ``[a] is the superclass of [b]. This statement is [MASK].'' Note here, $[P_{hierarchy}]$ is ``superclass''. Other three templates were based on $[P_{hierarchy}] \in$ {parent class, supertype, ancestor class}. And four more template instantiations predicated on the bottom-up taxonomy discovery were based on $[P_{hierarchy}] \in$ {subclass, child class, subtype, descendant class}. Thus eight experiments per template instantiation for the applicable LLM were run and the results from the best template were reported.

The prefix prompt template, similarly, reuses the cloze prompt template with the [MASK] token replaced with a blank or ``?'' symbol. It is $f_{p-prompt}^{B}(a, b) = [instruction] + f_{c-prompt}^{B}(a, b)$, with instruction ``Identify whether the following statement is true or false:''

\noindent{\textbf{Task C -} \underline{\textit{Non-Taxonomic Relation Extraction}}.} This task discovers non-taxonomic semantic heterarchical relations between types.

The cloze prompt template is $f_{c-prompt}^{C}(h, r, t) := [h]\;is\;[r]\;[t].\;This\;statement\;\\is\;[MASK].$ Where $h$ is a head type, $t$ is a tail type, and $r$ is a non-taxonomic relationship between $h$ and $r$. To support the discovery of a heterarchy that can consist of a 1-M relational cardinality, for a given relation, all possible type pairs of the ontology were created. The expected output for the [MASK] token was again true or false. Note, unlike in Task A and B, the given template was used as is and no variations of it were created.

Again, the prefix prompt template reuses the cloze prompt template as the other tasks, with instructions similar to task B. It is $f_{p-prompt}^{C}(h, r, t) = [instruction] + f_{c-prompt}^{C}(h, r, t)$

\section{LLMs4OL - Three Ontology Learning Tasks Evaluations}

\subsection{Evaluation Datasets - Ontological Knowledge Sources}

To comprehensively assess LLMs for the three OL tasks presented in the previous section, we cover a variety of ontological knowledge domain sources. Generally, across the tasks, four knowledge domains are represented, i.e. lexicosemantic -- WordNet~\cite{miller1995wordnet}, geographical -- GeoNames~\cite{geonames}, biomedicine -- Unified Medical Language System (UMLS)~\cite{umls} teased out as the National Cancer Institute (NCI)~\cite{NCI}, MEDCIN~\cite{MEDCIN}, and Systematized Nomenclature of Medicine -- Clinical Terms United States (SNOMEDCT\_US)~\cite{SNOMEDCT-US} subontologies, and content representations in the web -- schema.org~\cite{schemaorg}. Tasks A, B, and C applied only to UMLS. In other words, the ontology has a supporting knowledge base with terms that can be leveraged in the test prompts for term typing as Task A, taxonomic hierarchical relational prompts as Task B, and non-taxonomic heterarchical relational prompts as Task C. The GeoNames source came with a knowledge base of terms instantiated for types and taxonomic relations, therefore, was leveraged in the Task A and B as OL tests with LLMs of this work. The WordNet source could be leveraged only in Task A since it came with an instantiated collection of lexical terms for syntactic types. It was not applicable in the Tasks B and C for OL defined in this work since the semantic relations in WordNet are lexicosemantic, in other words, between terms directly and not their types. Finally, since the schema.org source offered only typed taxonomies as standardized downloads, it was leveraged only in the OL Task B of this work. In this case, we refrained from scraping the web for instantiations of the schema.org taxonomy. For all other ontological knowledge sources considered in this work that were relevant to Task A, the term instantiations were obtained directly from the source. This facilitates replicating our Task A dataset easily. Detailed information on the ontological knowledge sources per task with relevant dataset statistics are presented next. 

\noindent{\textbf{Task A Datasets.}} \autoref{taska:stats} shows statistical insights for the Task A dataset where we used terms from WordNet, GeoNames, and UMLS.  For WordNet we used the WN18RR data dump~\cite{wn18rr} that is derived from the original WordNet but released as a benchmark dataset with precreated train and test splits. Overall, it consists of 40,943 terms with 18 different relation types between the terms and four term types (noun, verb, adverb, adjective). We combined the original validation and test sets as a single test dataset. GeoNames comprises 680 categories of geographical locations, which are classified into 9 higher-level categories,  e.g. H for stream, lake, and sea, and R for road and railroad. UMLS contains almost three million concepts from various sources which are linked together by semantic relationships. UMLS is unique in that it is a greater semantic ontological network that subsumes other biomedical problem-domain restricted subontologies. We grounded the term typing task to the semantic spaces of three select subontological sources,i.e. NCI, MEDCIN, and SNOMEDCT\_US. 
% Table \ref{taska:stats} represents the dataset for Task A where we used entities with their types from WN18RR \cite{wn18rr}, GeoNames \cite{geonames}, and three different sources from UMLs \cite{umls}. 

\begin{table}[tb]
\centering
\caption{Task A term typing dataset counts across three core ontological knowledge sources, i.e. WordNet, GeoNames, and UMLS, where for Task A UMLS is represented only by the NCI, MEDCIN, and SNOMEDCT\_US subontological sources. The unique term types per source that defined Task A Ontology Learning is also provided.}\label{taska:stats}
\begin{tabular}{lrrrrr }
\hline
 Parameter &  \textbf{WordNet} & \textbf{GeoNames} & \textbf{NCI} & \textbf{MEDCIN} & \textbf{SNOMEDCT\_US} \\
\hline
\textit{Train Set Size} & 40,559 & 8,078,865 & 96,177 & 277,028 & 278,374 \\
\textit{Test Set Size}  &  9,470 & 702,510 & 24,045 & 69,258 & 69,594 \\
\textit{Types} &  4 & 680 & 125 & 87 & 125 \\
\hline
\end{tabular}
\end{table}

The train datasets were reserved for LLM fine-tuning. Among the 11 models, we selected the most promising one based on its zero-shot performance. The test datasets were used for evaluations in both zero-shot and fine-tuned settings.

\noindent{\textbf{Task B Datasets.}} From GeoNames, UMLS, and schema.org we obtained 689, 127, and 797 term types forming type taxonomies. 
Our test dataset was constructed as type pairs, where half represented the taxonomic hierarchy while the other half were not in a taxonomy. 
%Our testing was performed by building pairs between all types of which half actually constituted the taxonomic hierarchy while the other half were not in a taxonomy. 
This is based on the following formulations.
$$\forall(a \in T_n, b \in T_{n+1}) \longmapsto (aRb \wedge b\neg R a)$$
$$\forall(a\in T_n, b \in T_{n+1}, c \in T_{n+2}); (aRb \wedge bRc)  \longmapsto aRc$$
$$\forall(a\in T_n, b \in T_{n+1}, c \in T_{n+2}); (c\neg R b \wedge b\neg R a)  \longmapsto c\neg R a$$
Where $a$, $b$, and $c$ are types at different levels in the hierarchy. $T$ is a collection of types at a particular level in the taxonomy, where $n+2 > n+1 > n$ and $n$ is the root. The symbol $R$ represents ``$a$ is a super class of type $b$'' as a true taxonomic relation. Conversely, the $\neg R$ represents ``$b$ is a super class of type $a$'' as a false taxonomic relation. Furthermore, transitive taxonomic relations, $(aRb \wedge bRc)  \longmapsto aRc$, were also extracted as true relations, while their converse, i.e. $(c\neg R b \wedge b\neg R a)  \longmapsto c\neg R a$ were false relations.

%Moreover considering $a$, $b$, and $c$ from 3 consecutive levels as a set $X$, the $R$ relation on set $X$ has a transitive relation, which for all elements $\{a, b, c\} \in X$, Is-A relates $a$ to $b$ and $b$ to $c$. So, we can extract the fact "$a$ is a superior class to $c$" as a correct statement, since $aRb$, and $bRc$ (the equivalent happens for $\neg R$ as well). Table \ref{taskbc:stats} represents statistics for this task.

\noindent{\textbf{Task C Datasets.}} As alluded to earlier, Task C evaluations, i.e. non-taxonomic relations discovery, were relegated to the only available ontological knowledge source among those we considered i.e. UMLS. It reports 53 non-taxonomic relations across its 127 term types. The testing dataset comprised all pairs of types for each relation, where for any given relation some pairs are true while the rest are false candidates. Task B and Task C datasets' statistics are in \autoref{taskbc:stats}.

%as positive samples and using the negative sampling technique we created negative samples for the zero-shot classification task. 

%The negative sampling technique generates negative samples by corrupting a known positive $(h, r, t)$ triples in the semantic network, where $h$ and $t$ are‌ head and tail entity types and $r$ is a Non-Is-A relation. One of the interesting properties of the Non-Is-A relation in OL is the one-to-many property (1-M relations, where $M\in\mathbb{N}$) which means an entity type $h$ could have multiple properties or instances with $r$ relation; i.e. $h$ and $r$ could have many tail entity types such as $t_1, t_2, ...$ to form a $\{(h,r,t_1), (h,r,t_2), ... \}$, 1-M relations. Regarding this, negative sampler accounts for entities that co-occur with a relation $r$ to form a $(h, r, t^\prime)$ corrupted negative samples. A total number of 1-M relations with positive and negative stats are presented in \autoref{taskbc:stats}.

\begin{table}[tb]
\centering
\caption{Dataset statistics as counts per reported parameter for Task B type taxonomic hierarchy discovery and Task C type non-taxonomic heterarchy discovery across the pertinent ontological knowledge sources respectively per task.}\label{taskbc:stats}
\begin{tabular}{l l ccc}
\hline
 \textbf{Task} & Parameter &  \textbf{GeoNames} & \textbf{UMLS} & \textbf{schema.org} \\
\hline
\multirow{4}{*}{\textit{Task B}}&Types &  689 & 127 & 797 \\
&Levels &  2   & 3 & 6 \\
&Positive/Negative Samples  &  680/680 & 254/254 & 2,670/2,670 \\
% &Negative Samples &  680 & 254 & 2,670 \\
% &$\sum(positive + negative)$  &  1,360 & 508 & 5,340 \\
% \cline{2-5}
& \textit{Train/Test split} & 272/1,088 & 101/407&1,086/4,727 \\
% & \;\;\;Train Set Size & 272 & 101 & 1,068\\
% &‌ \;\;\;Test Set Size & 1,088 & 407 & 4,272 \\
\hline
\multirow{3}{*}{\textit{Task C}} & Non-Taxonomic Relations &  -  & 53 & - \\
%& 1-M samples & - & 683 &- \\
& Positive/Negative Samples  &  - & 5,641/1,896 & - \\
% & Negative Samples &  - & 1,896 & - \\
% &$\sum(positive + negative)$  &  - & 7,537  & - \\
% \cline{2-5}
& \textit{Train/Test Split} &  - & 1,507/6,030 & - \\
% & \;\;\;Train Set Size & - & 1,507 & - \\
% &‌ \;\;\;Test Set Size & - & 6,030 & - \\
\hline
\end{tabular}
\end{table}

%To fulfill our training and evaluation objectives, we split the task B and C sets such that only 20\% of the data was used for training, and 80\% was reserved for testing  (statistics are represented in the table \ref{taskbc:stats}). As for task A, we trained the model using a limited set of training set entities according to our model training strategy, during the training phase.

\subsection{Evaluation Models - Large Language Models (LLMs)}

As already introduced earlier, in this work, we comprehensively evaluate eight main types of domain-independent LLMs reported as state-of-the-art for different tasks in the community. They are: BERT~\cite{bert} as an encoder-only architecture, BLOOM~\cite{bloom}, LLaMA~\cite{llama}, GPT-3~\cite{gpt3}, GPT-3.5~\cite{chatgpt}, and GPT-4~\cite{gpt4} as decoder-only models, and finally BART~\cite{bart} and Flan-T5~\cite{flant5} as encoder-decoder models. Note these LLMs are released at varying parameter sizes. Thus qualified by the size in terms of parameters written in parenthesis, in all, we evaluate seven LLMs: 1. BERT-Large (340M), 2. BART-Large (400M), 3. Flan-T5-Large (780M), 4. Flan-T5-XL (3B), 5. BLOOM-1b7 (1.7B), 6. BLOOM-3b (3B), 7. GPT-3 (175B), 8. GPT-3.5 (174B), 9. LLaMA (7B), and GPT-4 ($>$1T). Additionally, we also test an eleventh biomedical domain-specific model PubMedBERT~\cite{pubmedbert}. 
%We evaluate five main types of LLMs representative of different neural architecture building blocks that are respectively reported as state-of-the-art for different tasks in the community. The tasks were briefly alluded to in the Introduction. Essentially, there are three main LLM architectures: encoder, decoder, and encoder-decoder. Representatively, we selected the following five main LLM types under the LLMs4OL three OL task paradigm evaluations: BERT~\cite{bert} as an encoder-only model, BLOOM~\cite{bloom} and GPT-3~\cite{gpt3} are decoder-only models, and finally BART~\cite{bart} and Flan-T5~\cite{flant5} as encoder-decoder models. Note these LLMs are released at varying parameter sizes. Thus qualified by the size in terms of parameters written in parenthesis, in all, we evaluate seven LLMs: 1. BERT-Large (340M), 2. BART-Large (400M), 3. Flan-T5-Large (780M), 4. Flan-T5-XL (3B), 5. BLOOM-1b7 (1.7B), 6. BLOOM-3b (3B), and 7. GPT-3 (175B).

In this work, since we propose the LLMs4OL paradigm for the first time, in a sense postulating OL as an emergent ability of LLMs, it is important for us to test different LLMs on the new task. Evaluating different LLMs supports: 1) Performance comparison - this allows us to identify which models are effective for OL, 2) Model improvement - toward OL one can identify areas where the models need improvement, and 3) Research advancement - with our results from testing and comparing different models, researchers interested in OL could potentially identify new areas of research and develop new techniques for improving LLMs.

\subsection{Evaluations}

\subsubsection{Metrics.}
\label{metrics}

Evaluations for Task A are reported as the mean average precision at k (MAP@K), where k = 1, since this metric was noted as being best suited to the task. Specifically, in our case, for term typing, MAP@1 measures the average precision of the top-1 ranked term types returned by an LLM for prompts initialized with terms from the evaluation set. And evaluations for Tasks B and C are reported in terms of the standard F1-score based on precision and recall.

\subsubsection{Results - Three Ontology Learning Tasks Zero-shot Evaluations.} The per task overall evaluations are reported in \autoref{taskabc-results-zeroshot}. %To our knowledge, the LLMs have not been trained explicitly for OL. Thus for the results reported in this section the LLMs are evaluated out-of-the-box, i.e. in a zero-shot setting. 
The three main rows of the table marked by alphabets A, B, and C correspond to term typing, type taxonomy discovery, and type non-taxonomic relational hetrarchy discovery results, respectively. The five subrows against Task A shows term typing results for WordNet, GeoNames, and the three UMLS subontologies, viz. NCI, SNOMEDCT\_US, and MEDCIN. The three subrows against Task B shows type taxonomy discovery results for GeoNames, UMLS, and schema.org, respectively. Task C evaluation results are provided only for UMLS. %The last seven columns of the results table correspond to the seven LLMs considered in the zero-shot three task OL evaluations. 
We first examine the results in the zero-shot setting, i.e. for LLMs evaluated out-of-the-box, w.r.t. three RQs.

\begin{table}[tb]
\centering
\caption{Zero-shot results across 11 LLMs and finetuned Flan-T5-Large and Flan-T5-XL LLMs results reported for ontology learning Task A i.e. term typing in MAP@1, and as F1-score for Task B i.e. type taxonomy discovery, and Task C i.e. type non-taxonomic relation extraction. The results are in percentages.}\label{taskabc-results-zeroshot}
\begin{tabular}{lcrrrrrrrrrrrp{0.3mm}rr}
\hline
& & \multicolumn{11}{c}{Zero-Shot Testing} & &\multicolumn{2}{c}{Finetuned} \\
\cline{3-13}
\cline{15-16}
\multirow{-8}{*}{\rotatebox{90}{\textbf{Task}}} & \multirow{-8}{*}{\rotatebox{90}{\textbf{Dataset}}} & \rotatebox{90}{\textbf{BERT-Large}} &  \rotatebox{90}{\textbf{PubMedBERT}} & \rotatebox{90}{\textbf{BART-Large}} &    \rotatebox{90}{\textbf{Flan-T5-Large}} & \rotatebox{90}{\textbf{Flan-T5-XL}} & \rotatebox{90}{\textbf{BLOOM-1b7}} &\rotatebox{90}{\textbf{BLOOM-3b}} & \rotatebox{90}{\textbf{GPT-3}} &  \rotatebox{90}{\textbf{GPT-3.5}} & \rotatebox{90}{\textbf{LLaMA-7B}} & \rotatebox{90}{\textbf{GPT-4}}& &\rotatebox{90}{\textbf{Flan-T5-Large$^{*}$}} & \rotatebox{90}{\textbf{Flan-T5-XL$^{*}$}}\\
\hline 
\multirow{5}{*}{\textit{A}}
&\textit{WordNet}   & 27.9 & - & 2.2  & 31.3 &  52.2  & 79.2 & 79.1 &  37.9 & \textbf{91.7} & 81.4 & 90.1 &  & 76.9 & \textbf{86.3} \\
&\textit{GeoNames} &  38.3 & - &  23.2&  13.2 &  33.8 & 28.5 &28.8 & 22.4 &35.0& 29.5 & \textbf{43.3} &  & 16.9 & 18.4\\
&\textit{NCI}      &  11.1 & 5.9 & 9.9  & 9.0  & 9.8  &  12.4 &   15.6 & 12.7 & 14.7 &7.7& \textbf{16.1} &  & 31.9 & \textbf{32.8}\\
&\textit{SNOMEDCT\_US} & 21.1 & 28.5 & 19.8 & 24.3 &  31.6  & 37.0  &    \textbf{37.7} & 24.4 & 25.0 & 13.8& 27.8 & &  33.4 & \textbf{43.4}\\
&\textit{MEDCIN}   & 8.7  & 15.6& 12.7 & 13.0 &  18.5   &  28.8  &  \textbf{29.8}  & 25.7 & 23.9 & 4.9 & 23.7 &  & 38.4 & \textbf{51.8}\\
\hline
\multirow{3}{*}{\textit{B}}
& \textit{GeoNames} & 54.5  & - & 55.4  & 59.6  & 52.4  & 36.7  &  48.3  & 53.2 & \textbf{67.8} &  33.5 &55.4 &  & 62.5 & 59.1 \\
& \textit{UMLS}     &  48.2  & 33.7& 49.9  & 55.3   & 64.3 &   38.3   &  37.5  & 51.6 &  70.4 & 32.3&\textbf{78.1} &  & 53.4 & \textbf{79.3}\\
& \textit{schema.org}   &  44.1  &- & 52.9  &  54.8    & 42.7 &   48.6   & 51.3  & 51.0 & \textbf{74.4} & 33.8&74.3 &  & 91.7 & \textbf{91.7}\\
\hline
\textit{C} & \textit{UMLS} & 40.1 & 42.7 & 42.4 & 46.0 & \textbf{49.5} &  43.1 &  42.7  & 38.8 & 37.5 & 20.3 & 41.3 &  & 49.1 & \textbf{53.1}\\
\hline
\end{tabular}
\end{table}
% \end{comment}

\noindent{\textbf{RQ1: How effective are LLMs for Task A, i.e. automated type discovery?}} We examine this question given the results in 5 subrows against the row A, i.e. corresponding to the various ontological datasets evaluated for Task A. Of the five ontological sources, the highest term typing results were achieved on the 4-typed WordNet at 91.7\% MAP@1 by GPT-3.5. This high performance can be attributed in part to the simple type space of WordNet with only 4 types. However, looking across the other LLMs evaluated on WordNet, in particular even GPT-3, scores in the range of 30\% MAP@1 seem to be the norm with a low of 2.2\% by BART-Large. Thus LLMs that report high scores on WordNet should be seen as more amenable to syntactic typing regardless of the WordNet simple type space. 
%Thus, instead, the high score of 76\% achieved by BLOOM on WordNet indicates that this LLM is more amenable to syntactic typing than the other evaluated LLMs. It has proven highly effective for syntactic typing in a zero-shot setting. 
Considering all the ontological sources, Geonames presents the most fine-grained types taxonomy of 680 types. Despite this, the best result obtained on this source is 39.4\% from GPT-4 with BERT-Large second at a close 38.3\%. This is better than the typing evaluations on the three biomedical datasets. Even the domain-specific PubMedBERT underperforms. In this regard, domain-independent models with large-scale parameters such a BLOOM (3B) are more amenable to this complex task. Since biomedicine entails deeper domain-specific semantics, we hypothesize better performance not just from domain-specific finetuning but also strategically for task-specific reasoning. 

The results overview is: 91.7\% WordNet by GPT-3.5 $>$ 39.4\% GeoNames by GPT-4 $>$ 37.7\% SNOMEDCT\_US by BLOOM-3b $>$ 29.8\% MEDCIN by BLOOM-3b $>$ 16.1\% NCI by GPT-4. 

\noindent{\textbf{RQ2: How effective are LLMs to recognize a type taxonomy i.e. the “is-a” hierarchy between types?}} We examine this question given the results in the 3 subrows against the main row B, i.e. corresponding to the three ontological sources evaluated for Task B. The highest result was achieved for UMLS by GPT-4 at 78.1\%. Of the open-source models, Flan-T5-XL achieved the best result at 64.3\%. Thus for the term taxonomy discovery LLMs on average have proven most effective in the zero-shot setting on the biomedical domain. 
%Intriguingly, the highest result was achieved at 64.25\% F1-score for UMLS by Flan-T5-XL. Thus for the term taxonomy discovery LLMs have proven most effective in a zero-shot prompt evaluation setting on the biomedical domain. 

The results overview is: 78.1\% UMLS by GPT-4 $>$ 74.4\% schema.org by GPT-3.5 $>$ 67.8\% GeoNames by GPT-3.5. Note the three GPT models were not open-sourced and thus we tested them with a paid subscription. For the open-source models, the results overview is: 64.3\% UMLS by Flan-T5-XL $>$ 59.6\% GeoNames by Flan-T5-XL $>$ 54.8\% schema.org by Flan-T5-Large. %These results order are also correlated with total taxonomic type pairs including both true pairs and false pairs which is as follows: 508 for UMLS, 1,360 for GeoNames, and 5,340 for schema.org.

\noindent{\textbf{RQ3: How effective are LLMs to discover non-taxonomic relations between types?}} We examine this question given the results in \autoref{taskabc-results-zeroshot} row for Task C, i.e. for UMLS. The best result achieved is 49.5\% by Flan-T5-XL. We consider this a fairly good result over a sizeable set of 7,537 type pairs that are in true non-taxonomic relations or are false pairs.

Finally, over all the three tasks considered under the LLMs4OL paradigm, term typing proved the hardest obtaining the lowest overall results for most of its ontological sources tested including the biomedical domain in particular. Additionally in our analysis, GPT, Flan-T5, and BLOOM variants showed improved scores with increase in parameters, respectively. This held true for the closed-sourced GPT models, i.e. GPT-3 (175B) and GPT-3.5 (175B) to GPT-4 ($>$1T) and the open-sourced models, i.e.  Flan-T5-Large (780M) to Flan-T5-XL (3B) and BLOOM from 1.7B to 3B. Thus it seems apparent that with an increased number of LLM parameters, we can expect an improvement in ontology learning.
%\item \textbf{Domain-specific zero-shot testing.} The domain-adjusted analysis of LLMs and the PubMedBERT results on biomedicine datasets revealed that it is performing similarly to the other LLMs and in a few cases even worse. This has led to an undervaluation of ontology learning tasks in biomedicine domains during LLM training.

%\vspace{-5mm}
\begin{figure}
\includegraphics[width=\textwidth]{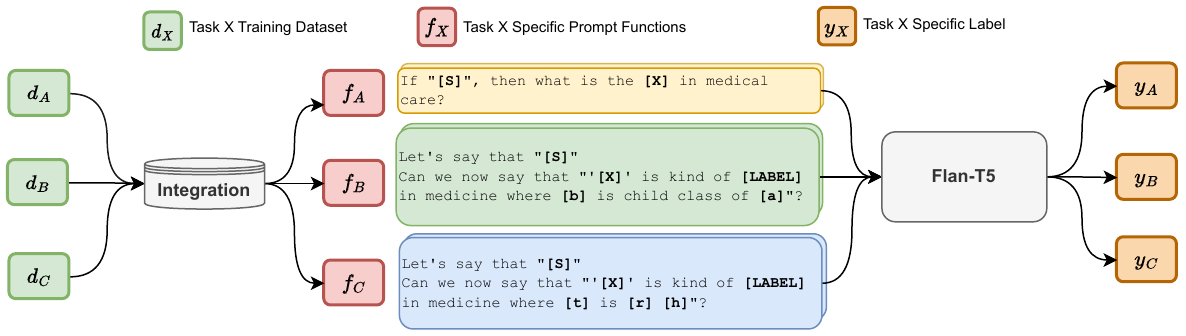}
\caption{An illustration of the LLM finetuning workflow on tasks for ontology learning.} \label{flant5-training}
%  We finetune Flan-T5 LM on three tasks and evaluate them on all three tasks using zero-shot testing. The prompt templates are examples from UMLS sources.
%\vspace{-1.1cm}
\end{figure} 
% \vspace{-1.1cm}
\subsubsection{Results - Three Ontology Learning Tasks Finetuned LLM Evaluations.}
Our zero-shot test results indicate that while LLMs seem promising for OL they would need task-specific finetuning to be a practically viable solution. To this end, we adopt the method of ``instruction tuning'' proposed as the \href{https://github.com/google-research/FLAN/blob/main/flan/templates.py}{FLAN collection} which is the only known systematically deconstructed, effective way to finetune LLMs~\cite{longpre2023flan}. For finetuning, we choose the Flan-T5 LMM for two reasons: 1) it is open-source: we intend to foster future research directions for models unhidden behind paywalls to aid in democratizing LLM research, and 2) it showed consistently good performance across all tasks. The finetuning instructions were instantiated from a small selection of eight samples of each knowledge source' reserved training set and fed in a finetuning workflow shown in \autoref{flant5-training}. The finetuned Flan models' results (see last two columns in \autoref{taskabc-results-zeroshot}) are significantly boosted across almost all tasks. For task A, we observed an average improvement of 25\% from zero-shot to the finetuned model for both Flan-T5 variants. Notably, SNOMEDCT\_US showed least improvement of 9\%, while the WordNet showed the most improvement of 45\%. For task B we marked an average improvement of 18\%, and for task C 3\%. Given an illustration of the results in \autoref{results-fig} shows that on average finetuned models, even with fewer parameters outperforms models with 1000x or more parameters across the three OL tasks. These insights appear crucial to expedite developmental research progress for practical tools for OL using LLMs which we plan to leverage in our future work.

\section{Conclusions and Future Directions} \label{conclusion}

Various initiatives benchmark LLM performance, revealing new task abilities \cite{srivastava2022beyond,wei2022emergent}. These benchmarks advance computer science's understanding of LLMs. We explore LLMs' potential for Ontology Learning \cite{gruber1995toward,maedche2001ontology} through our introduced conceptual framework, LLMs4OL. Extensive experiments on 11 LLMs across three OL tasks demonstrate the paradigm's proof of concept. Our \href{https://github.com/HamedBabaei/LLMs4OL}{codebase} facilitates replication and extension of methods for testing new LLMs. Our empirical results are promising to pave future work for OL.

Future research directions in the field of OL with LLMs can focus on several key areas. First, there is a need to enhance LLMs specifically for ontology learning tasks, exploring novel architectures and fine-tuning to capture ontological structures better. Second, expanding the evaluation to cover diverse knowledge domains beyond the ones examined in the current work would provide a broader understanding of LLMs' generalizability. Third, hybrid approaches that combine LLMs with traditional ontology learning techniques, such as lexico-syntactic pattern mining and clustering, could lead to more accurate and comprehensive ontologies. Fourth, further research can delve into the extraction of specific semantic relations, like part-whole relationships or causality, to enhance the expressiveness of learned ontologies. Standardizing evaluation metrics, creating benchmark datasets, exploring dynamic ontology evolution, and domain-specific learning are important directions. Additionally, integrating human-in-the-loop approaches with expert involvement would enhance ontology relevance and accuracy. Exploring these research directions will advance LLM-based ontology learning, enhancing knowledge acquisition and representation across domains.
%Pursuing these future research directions will advance ontology learning with LLMs, enabling more effective knowledge acquisition and representation across domains.

%In this paper we proposed LLMs4OL paradigm with three OL tasks for leveragint LLMs for ontology constructions. Our overarching aim with this work is to build a automated OL system based on LLMs. According to the \autoref{taskabc-results-zeroshot}, It is astonishing to see that few-shot learning on NCI source of UMLS for term typing resulted in outperformance in the other two sources, this opens another RQs for us "\textit{can we combine sources from different domains for training LLMs specific for term typing in the multiple domains?}". However, there is a limitation to this work, it is the weaker result of Flan-T5 models on the GeoNames dataset and we believe that since few-shot learning typically involves training a model on a small dataset with limited examples. Considering the test dataset size of GeoNames, the size of the few-shot dataset for training is too small/insufficient for the complexity of the task, which negatively impacts the performance of the model.  In summary, our proposed LLMs4OL paradigm shows promise for automating ontology construction, and the findings of this paper prompt further studies into the combination of different domain sources for training LLMs. Nevertheless, future work should address the limitations associated with few-shot learning with varying complexities to improve the overall performance of the models.

\paragraph*{Supplemental Material Statement:} Our LLM templates, detailed results, and codebase are publicly released as supplemental material on Github \url{https://github.com/HamedBabaei/LLMs4OL}.

%Source codes for LLMs4OL are available from Github. The few-shot learning hyperparameters setups, detailed results, and templates are attached with the submission on EasyChair and, if accepted, will be published on arXiv in an extended version of the paper. 

\begin{comment}
There are limitations to this work, the first limitation is the weaker result of Flan-T5 models on the GeoNames dataset and we believe that since few-shot learning typically involves training a model on a small dataset with limited examples. Considering the test dataset size of GeoNames, the size of the few-shot dataset for training is too small or insufficient for the complexity of the task, which negatively impacts the performance of the model. Since Flan-T5 benefits from extensive pretraining on the large corpus, providing it with a broader understanding of language patterns and contexts may boost the performance.

\hamed{another limitation is the answer set that we designed as a result of this we considered things simpler at the first step of LLMs4OL}

It is astonishing to see that few-shot learning on NCI concepts of UMLS for term typing resulted in outperformance in the other two sources, this opens another RQs for us: \textit{can we combine sources from different domains for training LLMs specific for term typing in the multiple domains?}.

\hamed{this is something that was in my mind for possible future work and if it is not appropriate please feel free to remove it from the conclusion}
% \section*{Acknowledgements}

\hamed{adding prompt templates in appendix}
\end{comment}

\section*{Author Contributions}

%Zhang San: Methodology, Conceptualization, Methodology, Software Priya Singh.: Data curation, Writing- Original draft preparation. Wang Wu: Visualization, Investigation. Jan Jansen: Supervision.: Ajay Kumar: Software, Validation.: Sun Qi: Writing- Reviewing and Editing,

Hamed Babaei Giglou: Conceptualization, Methodology, Software, Validation, Investigation, Resources, Data Curation, Writing - Original Draft, Visualization. Jennifer D'Souza: Conceptualization, Methodology, Investigation, Resources, Writing - Original Draft, Writing - Review \& Editing, Supervision, Project administration, Funding acquisition. S\"{o}ren Auer: Conceptualization, Methodology, Investigation, Resources, Review \& Editing, Supervision, Project administration, Funding acquisition.

\section*{Acknowledgements}

A 16-page final version of this paper has been accepted for publication in the research track of the 22nd International Semantic Web Conference (ISWC 2023). We thank the anonymous reviewers for their detailed and insightful comments on an earlier draft of the paper. This work was jointly supported by the German BMBF project SCINEXT (ID 01lS22070), DFG NFDI4DataScience (ID 460234259), and ERC ScienceGraph (ID 819536).

\bibliographystyle{splncs04}
\bibliography{mybibliography}

\begin{thebibliography}{10}
\providecommand{\url}[1]{\texttt{#1}}
\providecommand{\urlprefix}{URL }
\providecommand{\doi}[1]{https://doi.org/#1}

\bibitem{geonames}
Geonames geographical database (2023), \url{http://www.geonames.org/}

\bibitem{agirre2000enriching}
Agirre, E., Ansa, O., Hovy, E., Mart{\'\i}nez, D.: Enriching very large
  ontologies using the www. In: Proceedings of the First International
  Conference on Ontology Learning-Volume 31. pp. 25--30 (2000)

\bibitem{akkalyoncu-yilmaz-etal-2019-applying}
Akkalyoncu~Yilmaz, Z., Wang, S., Yang, W., Zhang, H., Lin, J.: Applying {BERT}
  to document retrieval with birch. In: Proceedings of the 2019 Conference on
  Empirical Methods in Natural Language Processing and the 9th International
  Joint Conference on Natural Language Processing (EMNLP-IJCNLP): System
  Demonstrations. pp. 19--24. Association for Computational Linguistics, Hong
  Kong, China (Nov 2019). \doi{10.18653/v1/D19-3004},
  \url{https://aclanthology.org/D19-3004}

\bibitem{alfonseca2002unsupervised}
Alfonseca, E., Manandhar, S.: An unsupervised method for general named entity
  recognition and automated concept discovery. In: Proceedings of the 1st
  international conference on general WordNet, Mysore, India. pp. 34--43 (2002)

\bibitem{amatriain2023transformer}
Amatriain, X.: Transformer models: an introduction and catalog. arXiv preprint
  arXiv:2302.07730  (2023)

\bibitem{r3-soa1}
Asim, M.N., Wasim, M., Khan, M.U.G., Mahmood, W., Abbasi, H.M.: A survey of
  ontology learning techniques and applications. Database  \textbf{2018},
  bay101 (2018)

\bibitem{umls}
Bodenreider, O.: {The Unified Medical Language System (UMLS): integrating
  biomedical terminology}. Nucleic Acids Research  \textbf{32}(suppl\_1),
  D267--D270 (01 2004). \doi{10.1093/nar/gkh061},
  \url{https://doi.org/10.1093/nar/gkh061}

\bibitem{bodenreider2004unified}
Bodenreider, O.: The unified medical language system (umls): integrating
  biomedical terminology. Nucleic acids research  \textbf{32}(suppl\_1),
  D267--D270 (2004)

\bibitem{gpt3}
Brown, T.B., Mann, B., Ryder, N., Subbiah, M., Kaplan, J., Dhariwal, P.,
  Neelakantan, A., Shyam, P., Sastry, G., Askell, A., Agarwal, S.,
  Herbert-Voss, A., Krueger, G., Henighan, T., Child, R., Ramesh, A., Ziegler,
  D.M., Wu, J., Winter, C., Hesse, C., Chen, M., Sigler, E., Litwin, M., Gray,
  S., Chess, B., Clark, J., Berner, C., McCandlish, S., Radford, A., Sutskever,
  I., Amodei, D.: Language models are few-shot learners (2020)

\bibitem{flant5}
Chung, H.W., Hou, L., Longpre, S., Zoph, B., Tay, Y., Fedus, W., Li, Y., Wang,
  X., Dehghani, M., Brahma, S., Webson, A., Gu, S.S., Dai, Z., Suzgun, M.,
  Chen, X., Chowdhery, A., Castro-Ros, A., Pellat, M., Robinson, K., Valter,
  D., Narang, S., Mishra, G., Yu, A., Zhao, V., Huang, Y., Dai, A., Yu, H.,
  Petrov, S., Chi, E.H., Dean, J., Devlin, J., Roberts, A., Zhou, D., Le, Q.V.,
  Wei, J.: Scaling instruction-finetuned language models (2022)

\bibitem{cui2021template}
Cui, L., Wu, Y., Liu, J., Yang, S., Zhang, Y.: Template-based named entity
  recognition using bart. arXiv preprint arXiv:2106.01760  (2021)

\bibitem{dai2019transformerxl}
Dai, Z., Yang, Z., Yang, Y., Carbonell, J., Le, Q.V., Salakhutdinov, R.:
  Transformer-xl: Attentive language models beyond a fixed-length context
  (2019)

\bibitem{dalvi2022discovering}
Dalvi, F., Khan, A.R., Alam, F., Durrani, N., Xu, J., Sajjad, H.: Discovering
  latent concepts learned in {BERT}. In: International Conference on Learning
  Representations (2022), \url{https://openreview.net/forum?id=POTMtpYI1xH}

\bibitem{wn18rr}
Dettmers, T., Pasquale, M., Pontus, S., Riedel, S.: Convolutional 2d knowledge
  graph embeddings. In: Proceedings of the 32th AAAI Conference on Artificial
  Intelligence. pp. 1811--1818 (February 2018),
  \url{https://arxiv.org/abs/1707.01476}

\bibitem{bert}
Devlin, J., Chang, M.W., Lee, K., Toutanova, K.: Bert: Pre-training of deep
  bidirectional transformers for language understanding (2019)

\bibitem{dopazo1997phylogenetic}
Dopazo, J., Carazo, J.M.: Phylogenetic reconstruction using an unsupervised
  growing neural network that adopts the topology of a phylogenetic tree.
  Journal of molecular evolution  \textbf{44}(2),  226--233 (1997)

\bibitem{gruber1995toward}
Gruber, T.R.: Toward principles for the design of ontologies used for knowledge
  sharing? International journal of human-computer studies  \textbf{43}(5-6),
  907--928 (1995)

\bibitem{pubmedbert}
Gu, Y., Tinn, R., Cheng, H., Lucas, M., Usuyama, N., Liu, X., Naumann, T., Gao,
  J., Poon, H.: Domain-specific language model pretraining for biomedical
  natural language processing. ACM Transactions on Computing for Healthcare
  (HEALTH)  \textbf{3}(1),  1--23 (2021)

\bibitem{guha2016schema}
Guha, R.V., Brickley, D., Macbeth, S.: Schema. org: evolution of structured
  data on the web. Communications of the ACM  \textbf{59}(2),  44--51 (2016)

\bibitem{hahn2001joint}
Hahn, U., Mark{\'o}, K.G.: Joint knowledge capture for grammars and ontologies.
  In: Proceedings of the 1st international conference on Knowledge capture. pp.
  68--75 (2001)

\bibitem{hamp1997germanet}
Hamp, B., Feldweg, H.: Germanet-a lexical-semantic net for german. In:
  Automatic information extraction and building of lexical semantic resources
  for NLP applications (1997)

\bibitem{hearst1998automated}
Hearst, M.A.: Automated discovery of wordnet relations. WordNet: an electronic
  lexical database  \textbf{2} (1998)

\bibitem{hwang1999incompletely}
Hwang, C.H.: Incompletely and imprecisely speaking: using dynamic ontologies
  for representing and retrieving information. In: KRDB. vol.~21, pp. 14--20.
  Citeseer (1999)

\bibitem{jiang-etal-2020-know}
Jiang, Z., Xu, F.F., Araki, J., Neubig, G.: How can we know what language
  models know? Transactions of the Association for Computational Linguistics
  \textbf{8},  423--438 (2020). \doi{10.1162/tacl\_a\_00324},
  \url{https://aclanthology.org/2020.tacl-1.28}

\bibitem{khan2002ontology}
Khan, L., Luo, F.: Ontology construction for information selection. In: 14th
  IEEE International Conference on Tools with Artificial Intelligence,
  2002.(ICTAI 2002). Proceedings. pp. 122--127. IEEE (2002)

\bibitem{khot2023decomposed}
Khot, T., Trivedi, H., Finlayson, M., Fu, Y., Richardson, K., Clark, P.,
  Sabharwal, A.: Decomposed prompting: A modular approach for solving complex
  tasks (2023)

\bibitem{kietz2000method}
Kietz, J.U., Maedche, A., Volz, R.: A method for semi-automatic ontology
  acquisition from a corporate intranet. In: EKAW-2000 Workshop “Ontologies
  and Text”, Juan-Les-Pins, France, October 2000 (2000)

\bibitem{kojima2023large}
Kojima, T., Gu, S.S., Reid, M., Matsuo, Y., Iwasawa, Y.: Large language models
  are zero-shot reasoners (2023)

\bibitem{konys2019knowledge}
Konys, A.: Knowledge repository of ontology learning tools from text. Procedia
  Computer Science  \textbf{159},  1614--1628 (2019)

\bibitem{lester2021power}
Lester, B., Al-Rfou, R., Constant, N.: The power of scale for
  parameter-efficient prompt tuning. arXiv preprint arXiv:2104.08691  (2021)

\bibitem{levy-etal-2017-zero}
Levy, O., Seo, M., Choi, E., Zettlemoyer, L.: Zero-shot relation extraction via
  reading comprehension. In: Proceedings of the 21st Conference on
  Computational Natural Language Learning ({C}o{NLL} 2017). pp. 333--342.
  Association for Computational Linguistics, Vancouver, Canada (Aug 2017).
  \doi{10.18653/v1/K17-1034}, \url{https://aclanthology.org/K17-1034}

\bibitem{bart}
Lewis, M., Liu, Y., Goyal, N., Ghazvininejad, M., Mohamed, A., Levy, O.,
  Stoyanov, V., Zettlemoyer, L.: Bart: Denoising sequence-to-sequence
  pre-training for natural language generation, translation, and comprehension
  (2019)

\bibitem{li2021prefix}
Li, X.L., Liang, P.: Prefix-tuning: Optimizing continuous prompts for
  generation. arXiv preprint arXiv:2101.00190  (2021)

\bibitem{prompting}
Liu, P., Yuan, W., Fu, J., Jiang, Z., Hayashi, H., Neubig, G.: Pre-train,
  prompt, and predict: A systematic survey of prompting methods in natural
  language processing. ACM Comput. Surv.  \textbf{55}(9) (jan 2023).
  \doi{10.1145/3560815}, \url{https://doi.org/10.1145/3560815}

\bibitem{longpre2023flan}
Longpre, S., Hou, L., Vu, T., Webson, A., Chung, H.W., Tay, Y., Zhou, D., Le,
  Q.V., Zoph, B., Wei, J., et~al.: The flan collection: Designing data and
  methods for effective instruction tuning. arXiv preprint arXiv:2301.13688
  (2023)

\bibitem{lonsdale2002peppering}
Lonsdale, D., Ding, Y., Embley, D.W., Melby, A.: Peppering knowledge sources
  with salt: Boosting conceptual content for ontology generation. In:
  Proceedings of the AAAI Workshop on Semantic Web Meets Language Resources,
  Edmonton, Alberta, Canada (2002)

\bibitem{r3-soa2}
Lourdusamy, R., Abraham, S.: A survey on methods of ontology learning from
  text. In: Intelligent Computing Paradigm and Cutting-edge Technologies:
  Proceedings of the First International Conference on Innovative Computing and
  Cutting-edge Technologies (ICICCT 2019), Istanbul, Turkey, October 30-31,
  2019 1. pp. 113--123. Springer (2020)

\bibitem{maedche2001ontology}
Maedche, A., Staab, S.: Ontology learning for the semantic web. IEEE
  Intelligent systems  \textbf{16}(2),  72--79 (2001)

\bibitem{MEDCIN}
{Medicomp Systems}: {MEDCIN} (January 2023), \url{https://medicomp.com}

\bibitem{miller1995wordnet}
Miller, G.A.: Wordnet: a lexical database for english. Communications of the
  ACM  \textbf{38}(11),  39--41 (1995)

\bibitem{missikoff2002usable}
Missikoff, M., Navigli, R., Velardi, P.: The usable ontology: An environment
  for building and assessing a domain ontology. In: The Semantic Web—ISWC
  2002: First International Semantic Web Conference Sardinia, Italy, June
  9--12, 2002 Proceedings. pp. 39--53. Springer (2002)

\bibitem{moldovan2001interactive}
Moldovan, D.I., GiRJU, R.C.: An interactive tool for the rapid development of
  knowledge bases. International Journal on Artificial Intelligence Tools
  \textbf{10}(01n02),  65--86 (2001)

\bibitem{NCI}
{National Cancer Institute, National Institutes of Health}: {NCI Thesaurus}
  (September 2022), \url{http://ncit.nci.nih.gov}

\bibitem{noy2001ontology}
Noy, N.F., McGuinness, D.L., et~al.: Ontology development 101: A guide to
  creating your first ontology (2001)

\bibitem{chatgpt}
OpenAI: Chatgpt. \url{https://openai.com/chat-gpt/} (2023), accessed May 5,
  2023

\bibitem{gpt4}
OpenAI: Gpt-4 technical report (2023)

\bibitem{schemaorg}
Patel-Schneider, P.F.: Analyzing schema.org. In: Mika, P., Tudorache, T.,
  Bernstein, A., Welty, C., Knoblock, C., Vrande{\v{c}}i{\'{c}}, D., Groth, P.,
  Noy, N., Janowicz, K., Goble, C. (eds.) The Semantic Web -- ISWC 2014. pp.
  261--276. Springer International Publishing, Cham (2014)

\bibitem{peters-etal-2018-deep}
Peters, M.E., Neumann, M., Iyyer, M., Gardner, M., Clark, C., Lee, K.,
  Zettlemoyer, L.: Deep contextualized word representations. In: Proceedings of
  the 2018 Conference of the North {A}merican Chapter of the Association for
  Computational Linguistics: Human Language Technologies, Volume 1 (Long
  Papers). pp. 2227--2237. Association for Computational Linguistics, New
  Orleans, Louisiana (Jun 2018). \doi{10.18653/v1/N18-1202},
  \url{https://aclanthology.org/N18-1202}

\bibitem{petroni2020how}
Petroni, F., Lewis, P., Piktus, A., Rockt{\"a}schel, T., Wu, Y., Miller, A.H.,
  Riedel, S.: How context affects language models' factual predictions. In:
  Automated Knowledge Base Construction (2020),
  \url{https://openreview.net/forum?id=025X0zPfn}

\bibitem{petroni2019language}
Petroni, F., Rockt{\"a}schel, T., Lewis, P., Bakhtin, A., Wu, Y., Miller, A.H.,
  Riedel, S.: Language models as knowledge bases? arXiv preprint
  arXiv:1909.01066  (2019)

\bibitem{lms-as-kb}
Petroni, F., Rockt{\"a}schel, T., Riedel, S., Lewis, P., Bakhtin, A., Wu, Y.,
  Miller, A.: Language models as knowledge bases? In: Proceedings of the 2019
  Conference on Empirical Methods in Natural Language Processing and the 9th
  International Joint Conference on Natural Language Processing (EMNLP-IJCNLP).
  Association for Computational Linguistics (2019)

\bibitem{rebele2016yago}
Rebele, T., Suchanek, F., Hoffart, J., Biega, J., Kuzey, E., Weikum, G.: Yago:
  A multilingual knowledge base from wikipedia, wordnet, and geonames. In: The
  Semantic Web--ISWC 2016: 15th International Semantic Web Conference, Kobe,
  Japan, October 17--21, 2016, Proceedings, Part II 15. pp. 177--185. Springer
  (2016)

\bibitem{roux2000ontology}
Roux, C., Proux, D., Rechenmann, F., Julliard, L.: An ontology enrichment
  method for a pragmatic information extraction system gathering data on
  genetic interactions. In: ECAI Workshop on Ontology Learning (2000)

\bibitem{sajjad2022analyzing}
Sajjad, H., Durrani, N., Dalvi, F., Alam, F., Khan, A.R., Xu, J.: Analyzing
  encoded concepts in transformer language models (2022)

\bibitem{bloom}
Scao, T.L., Fan, A., Akiki, C., Pavlick, E., Ili{\'c}, S., Hesslow, D.,
  Castagn{\'e}, R., Luccioni, A.S., Yvon, F., Gall{\'e}, M., et~al.: Bloom: A
  176b-parameter open-access multilingual language model. arXiv preprint
  arXiv:2211.05100  (2022)

\bibitem{SNOMEDCT-US}
{SNOMED International}: {US Edition of SNOMED CT} (March 2023),
  \url{https://www.nlm.nih.gov/healthit/snomedct/us_edition.html}

\bibitem{srivastava2022beyond}
Srivastava, A., Rastogi, A., Rao, A., Shoeb, A.A.M., Abid, A., Fisch, A.,
  Brown, A.R., Santoro, A., Gupta, A., Garriga-Alonso, A., et~al.: Beyond the
  imitation game: Quantifying and extrapolating the capabilities of language
  models. arXiv preprint arXiv:2206.04615  (2022)

\bibitem{llama}
Touvron, H., Lavril, T., Izacard, G., Martinet, X., Lachaux, M.A., Lacroix, T.,
  Rozi{\`e}re, B., Goyal, N., Hambro, E., Azhar, F., et~al.: Llama: Open and
  efficient foundation language models. arXiv preprint arXiv:2302.13971  (2023)

\bibitem{wagner2000enriching}
Wagner, A.: Enriching a lexical semantic net with selectional preferences by
  means of statistical corpus analysis. In: ECAI Workshop on Ontology Learning.
  vol.~61. Citeseer (2000)

\bibitem{wkatrobski2020ontology}
W{\c{a}}tr{\'o}bski, J.: Ontology learning methods from text-an extensive
  knowledge-based approach. Procedia Computer Science  \textbf{176},
  3356--3368 (2020)

\bibitem{wei2022finetuned}
Wei, J., Bosma, M., Zhao, V., Guu, K., Yu, A.W., Lester, B., Du, N., Dai, A.M.,
  Le, Q.V.: Finetuned language models are zero-shot learners. In: International
  Conference on Learning Representations (2022),
  \url{https://openreview.net/forum?id=gEZrGCozdqR}

\bibitem{wei2022emergent}
Wei, J., Tay, Y., Bommasani, R., Raffel, C., Zoph, B., Borgeaud, S., Yogatama,
  D., Bosma, M., Zhou, D., Metzler, D., et~al.: Emergent abilities of large
  language models. arXiv preprint arXiv:2206.07682  (2022)

\bibitem{NEURIPS20229d560961}
Wei, J., Wang, X., Schuurmans, D., Bosma, M., ichter, b., Xia, F., Chi, E., Le,
  Q.V., Zhou, D.: Chain-of-thought prompting elicits reasoning in large
  language models. In: Koyejo, S., Mohamed, S., Agarwal, A., Belgrave, D., Cho,
  K., Oh, A. (eds.) Advances in Neural Information Processing Systems. vol.~35,
  pp. 24824--24837. Curran Associates, Inc. (2022),
  \url{https://proceedings.neurips.cc/paper_files/paper/2022/file/9d5609613524ecf4f15af0f7b31abca4-Paper-Conference.pdf}

\bibitem{weibel2000dublin}
Weibel, S.L., Koch, T.: The dublin core metadata initiative. D-lib magazine
  \textbf{6}(12),  1082--9873 (2000)

\bibitem{xu2002domain}
Xu, F., Kurz, D., Piskorski, J., Schmeier, S.: A domain adaptive approach to
  automatic acquisition of domain relevant terms and their relations with
  bootstrapping. In: LREC (2002)

\bibitem{yang2019simple}
Yang, W., Zhang, H., Lin, J.: Simple applications of bert for ad hoc document
  retrieval (2019)

\bibitem{yao2023tree}
Yao, S., Yu, D., Zhao, J., Shafran, I., Griffiths, T.L., Cao, Y., Narasimhan,
  K.: Tree of thoughts: Deliberate problem solving with large language models
  (2023)

\end{thebibliography}

\appendix 
\section{Apendix}

\subsection{Task A}

\textbf{Prompt Templates for WordNet Dataset.} Templates for WordNet in zero-shot testing for Task A are presented in \autoref{task-a-template-wordnet}. \\

\noindent \textbf{Prompt Templates for GeoNames Dataset.} Templates for GeoNames in zero-shot testing for Task A are presented in \autoref{task-a-template-geonames}. As a sentence $S$, we used $[L]\;is\;a\;place\;in\;[COUNTRY].$ template. \\

\noindent \textbf{Prompt Templates for UMLS Dataset.} Templates for UMLS sources (NCI, MEDCIN, and SNOMEDCT\_US) in zero-shot testing for Task A are presented in \autoref{task-a-template-umls}.\\

\begin{table}[]
    \centering
    \caption{The WordNet zero-shot testing prompt templates for task A. $L$ represents lexical entries, $S$ represents sentence containing $L$. In the BERT/BART‌ LLMs, for BART the $[MASK]$ is being replaced by $<mask>"$.}
    \label{task-a-template-wordnet}
    \begin{tabular}{l   l}
        \hline
        \textbf{LLMs} & \textit{Prompt Templates} \\
        \hline
        \multirow{8}{*}{\textit{BERT/BART} }
         & [S]. [L] POS is a $[MASK]$ . \\
         & [S]. [L] part of speech is a $[MASK]$ . \\
         & [S]. '[L]' POS is a $[MASK]$ .\\
         & [S]. '[L]' part of speech is a $[MASK]$ .\\
         & [L] POS is a $[MASK]$ .\\
         & [L] part of speech is a $[MASK]$ .\\
         & '[L]' POS is a $[MASK]$ .\\
         & '[L]' part of speech is a $[MASK]$ .\\
        \hline
        % \multirow{8}{*}{\textit{BART} }
        %  & [S]. [L] POS is a $<mask>$ . \\
        %  & [S]. [L] part of speech is a $<mask>$ . \\
        %  & [S]. '[L]' POS is a $<mask>$ . \\
        %  & [S]. '[L]' part of speech is a $<mask>$ .\\
        %  & [L] POS is a $<mask>$ .\\
        %  & [L] part of speech is a $<mask>$ .\\
        %  & '[L]' POS is a $<mask>$ .\\
        %  & '[L]' part of speech is a $<mask>$ .\\
        % \hline
         \multirow{8}{*}{\textit{Flan-T5} }
         & [S]. [L] POS is a ?\\
         & [S]. [L] part of speech is a ? \\
         & [S]. '[L]' POS is a ? \\
         & [S]. '[L]' part of speech is a ?\\
         & [L] POS is a ?\\
         & [L] part of speech is a ?\\
         & '[L]' POS is a ?\\
         & '[L]' part of speech is a ?\\
        \hline
        \multirow{16}{*}{\textit{BLOOM/GPT-3} }
         &  \multirow{2}{*}{\parbox{9cm}{Perform a sentence completion on the following sentence: \textbackslash n
                          Sentence: [S]. [L] POS is a}}\\
           & \\
         \cline{2-2}
         &  \multirow{2}{*}{\parbox{9cm}{Perform a sentence completion on the following sentence: \textbackslash n
                         Sentence: [S]. [L] part of speech is a}}\\
          & \\
         \cline{2-2}
         &  \multirow{2}{*}{\parbox{9cm}{Perform a sentence completion on the following sentence: \textbackslash n
                         Sentence: [S]. '[L]' POS is a }}\\
          & \\
         \cline{2-2}
         &  \multirow{2}{*}{\parbox{9cm}{Perform a sentence completion on the following sentence: \textbackslash n
                         Sentence: [S]. '[L]' part of speech is a }}\\
          & \\
         \cline{2-2}
         &  \multirow{2}{*}{\parbox{9cm}{Perform a sentence completion on the following sentence: \textbackslash n
                         Sentence: [L] POS is a}}\\
          & \\
         \cline{2-2}
         &  \multirow{2}{*}{\parbox{9cm}{Perform a sentence completion on the following sentence: \textbackslash n
                         Sentence: [L] part of speech is a}}\\
          & \\
         \cline{2-2}
         &  \multirow{2}{*}{\parbox{9cm}{Perform a sentence completion on the following sentence: \textbackslash n
                         Sentence: '[L]' POS is a }}\\
          & \\
         \cline{2-2}
         &  \multirow{2}{*}{\parbox{9cm}{Perform a sentence completion on the following sentence: \textbackslash n
                         Sentence: '[L]' part of speech is a}}\\
         & \\
        \hline

        \multirow{16}{*}{\textit{LLaMA} }
         &  \multirow{2}{*}{\parbox{9cm}{Perform a sentence completion on the following sentence: 
                          [S]. [L] POS is a  \_\_\_. \textbackslash n The answer is}}\\
           & \\
         \cline{2-2}
         &  \multirow{2}{*}{\parbox{9cm}{Perform a sentence completion on the following sentence:
                         [S]. [L] part of speech is a \_\_\_. \textbackslash n The answer is}}\\
          & \\
         \cline{2-2}
         &  \multirow{2}{*}{\parbox{9cm}{Perform a sentence completion on the following sentence:
                         [S]. '[L]' POS is a \_\_\_. \textbackslash n The answer is}}\\
          & \\
         \cline{2-2}
         &  \multirow{2}{*}{\parbox{9cm}{Perform a sentence completion on the following sentence:
                         [S]. '[L]' part of speech is a \_\_\_. \textbackslash n The answer is}}\\
          & \\
         \cline{2-2}
         &  \multirow{2}{*}{\parbox{9cm}{Perform a sentence completion on the following sentence:
                         [L] POS is a \_\_\_. \textbackslash n The answer is}}\\
          & \\
         \cline{2-2}
         &  \multirow{2}{*}{\parbox{9cm}{Perform a sentence completion on the following sentence:
                         [L] part of speech is a \_\_\_. \textbackslash n The answer is}}\\
          & \\
         \cline{2-2}
         &  \multirow{2}{*}{\parbox{9cm}{Perform a sentence completion on the following sentence:
                         '[L]' POS is a \_\_\_. \textbackslash n The answer is}}\\
          & \\
         \cline{2-2}
         &  \multirow{2}{*}{\parbox{9cm}{Perform a sentence completion on the following sentence:
                         '[L]' part of speech is a \_\_\_. \textbackslash n The answer is}}\\
         & \\
        \hline
    \end{tabular}
\end{table}

\begin{table}[]
    \centering
    \caption{The GeoNames zero-shot testing prompt templates for task A. $L$ represents lexical entries, $S$ represents sentence containing $L$. In the BERT/BART‌ LLMs, for BART the $[MASK]$ is being replaced by $<mask>"$.}
    \label{task-a-template-geonames}
    \begin{tabular}{l   l}
        \hline
        \textbf{LLMs} & \textbf{Prompt Templates} \\
        \hline
        \multirow{8}{*}{\textit{BERT/BART}} 
         & [S]. [L] is a $[MASK]$ . \\
         & [S]. [L] geographically is a $[MASK]$ . \\
         & [S]. '[L]' is a $[MASK]$ . \\
         & [S]. '[L]' geographically is a $[MASK]$ .\\
         & [L] is a $[MASK]$ .\\
         & [L] geographically is a $[MASK]$ .\\
         & '[L]' is a $[MASK]$ .\\
         & '[L]' geographically is a $[MASK]$ .\\
        % \hline
        % \multirow{8}{*}{\textit{BART}} 
        %  & [S]. [L] is a $<mask>$ . \\
        %  & [S]. [L] geographically is a $<mask>$ . \\
        %  & [S]. '[L]' is a $<mask>$ . \\
        %  & [S]. '[L]' geographically is a $<mask>$ .\\
        %  & [L] is a $<mask>$  .\\
        %  & [L] geographically is a $<mask>$  .\\
        %  & '[L]' is a $<mask>$ .\\
        %  & '[L]' geographically is a $<mask>$  .\\
        \hline
         \multirow{8}{*}{\textit{Flan-T5}} 
         & [S]. [L] is a ?\\
         & [S]. [L] geographically is a  ?\\
         & [S]. '[L]' is a  ?\\
         & [S]. '[L]' geographically is a ?\\
         & [L] is a ?\\
         & [L] geographically is a ?\\
         & '[L]' is a ?\\
         & '[L]' geographically is a ?\\
        \hline
        \multirow{16}{*}{\textit{BLOOM/GPT-3}} 
         &  \multirow{2}{*}{\parbox{9cm}{Perform a sentence completion on the following sentence: \textbackslash n
                          Sentence: [S]. [L] is a}}\\
         & \\
         \cline{2-2}
         &  \multirow{2}{*}{\parbox{9cm}{Perform a sentence completion on the following sentence: \textbackslash n
                         Sentence: [S]. [L] geographically is a}}\\
         & \\
         \cline{2-2}
         &  \multirow{2}{*}{\parbox{9cm}{Perform a sentence completion on the following sentence: \textbackslash n
                         Sentence: [S]. '[L]' is a }}\\
         & \\
         \cline{2-2}
         &  \multirow{2}{*}{\parbox{9cm}{Perform a sentence completion on the following sentence: \textbackslash n
                          Sentence: [S]. '[L]' geographically is a }}\\
         & \\
         \cline{2-2}
         &  \multirow{2}{*}{\parbox{9cm}{Perform a sentence completion on the following sentence: \textbackslash n
                         Sentence: [L] is a}}\\
         & \\
         \cline{2-2}
         &  \multirow{2}{*}{\parbox{9cm}{Perform a sentence completion on the following sentence: \textbackslash n
                         Sentence: [L] geographically is a}}\\
         & \\
         \cline{2-2}
         &  \multirow{2}{*}{\parbox{9cm}{Perform a sentence completion on the following sentence: \textbackslash n
                         Sentence: '[L]' is a }}\\
         & \\
         \cline{2-2}
         &  \multirow{2}{*}{\parbox{9cm}{Perform a sentence completion on the following sentence: \textbackslash n
                         Sentence: '[L]' geographically is a}}\\
         & \\
        \hline
        \multirow{16}{*}{\textit{LLaMA}} 
         &  \multirow{2}{*}{\parbox{9cm}{Perform a sentence completion on the following sentence:
                          [S]. [L] is a \_\_\_. \textbackslash n The answer is}}\\
         & \\
         \cline{2-2}
         &  \multirow{2}{*}{\parbox{9cm}{Perform a sentence completion on the following sentence:
                         [S]. [L] geographically is a \_\_\_. \textbackslash n The answer is}}\\
         & \\
         \cline{2-2}
         &  \multirow{2}{*}{\parbox{9cm}{Perform a sentence completion on the following sentence: 
                         [S]. '[L]' is a \_\_\_. \textbackslash n The answer is }}\\
         & \\
         \cline{2-2}
         &  \multirow{2}{*}{\parbox{9cm}{Perform a sentence completion on the following sentence: \textbackslash n
                          [S]. '[L]' geographically is a \_\_\_. \textbackslash n The answer is }}\\
         & \\
         \cline{2-2}
         &  \multirow{2}{*}{\parbox{9cm}{Perform a sentence completion on the following sentence: \textbackslash n
                         [L] is a \_\_\_. \textbackslash n The answer is}}\\
         & \\
         \cline{2-2}
         &  \multirow{2}{*}{\parbox{9cm}{Perform a sentence completion on the following sentence: \textbackslash n
                         [L] geographically is a \_\_\_. \textbackslash n The answer is}}\\
         & \\
         \cline{2-2}
         &  \multirow{2}{*}{\parbox{9cm}{Perform a sentence completion on the following sentence: \textbackslash n
                         '[L]' is a \_\_\_. \textbackslash n The answer is }}\\
         & \\
         \cline{2-2}
         &  \multirow{2}{*}{\parbox{9cm}{Perform a sentence completion on the following sentence: \textbackslash n
                         '[L]' geographically is a \_\_\_. \textbackslash n The answer is}}\\
         & \\
        \hline
    \end{tabular}
\end{table}

\begin{table}[]
    \centering
    \caption{The UMLS zero-shot testing prompt templates for task A. $L$ represents lexical entries, $S$ represents sentence containing $L$. In the BERT/BART‌ LLMs, for BART the $[MASK]$ is being replaced by $<mask>"$.}
    \label{task-a-template-umls}
    \begin{tabular}{l   l}
        \hline
        \textbf{LLMs} & \textbf{Prompt Templates} \\
        \hline
        \multirow{8}{*}{\textit{BERT/BART} }
         & [S]. [L] in medicine is a $[MASK]$ . \\
         & [S]. [L] in biomedicine is a $[MASK]$ . \\
         & [S]. '[L]' in medicine is a $[MASK]$ . \\
         & [S]. '[L]' in biomedicine is a $[MASK]$ .\\
         & [L] in medicine is a $[MASK]$ .\\
         & [L] in biomedicine is a $[MASK]$ .\\
         & '[L]' is a $[MASK]$ .\\
         & '[L]' in biomedicine is a $[MASK]$ .\\
        % \hline
        % \multirow{8}{*}{\textit{BART} }
        %  & [S]. [L] in medicine is a $<mask>$ . \\
        %  & [S]. [L] in biomedicine is a $<mask>$  . \\
        %  & [S]. '[L]' in medicine is a $<mask>$  . \\
        %  & [S]. '[L]' in biomedicine is a $<mask>$ .\\
        %  & [L] in medicine is a $<mask>$  .\\
        %  & [L] in biomedicine is a $<mask>$  .\\
        %  & '[L]' in medicine is a $<mask>$  .\\
        %  & '[L]' in biomedicine is a $<mask>$ .\\
        \hline
         \multirow{8}{*}{\textit{Flan-T5}}
         & [S]. [L] in medicine is a ?\\
         & [S]. [L] in biomedicine is a  ?\\
         & [S]. '[L]' in medicine is a ? \\
         & [S]. '[L]' in biomedicine is a ?\\
         & [L] in medicine is a ?\\
         & [L] in biomedicine is a ?\\
         & '[L]' in medicine is a ?\\
         & '[L]' in biomedicine is a ?\\
        \hline
        \multirow{16}{*}{\textit{BLOOM/GPT-3}} 
         &  \multirow{2}{*}{\parbox{9cm}{Perform a sentence completion on the following sentence: \textbackslash n
                          Sentence: [S]. [L] in medicine is a}}\\
         & \\
         \cline{2-2}
         &  \multirow{2}{*}{\parbox{9cm}{Perform a sentence completion on the following sentence: \textbackslash n
                         Sentence: [S]. [L] in biomedicine is a}}\\
         & \\
         \cline{2-2}
         &  \multirow{2}{*}{\parbox{9cm}{Perform a sentence completion on the following sentence: \textbackslash n
                         Sentence: [S]. '[L]' in medicine is a }}\\
         & \\
         \cline{2-2}
         &  \multirow{2}{*}{\parbox{9cm}{Perform a sentence completion on the following sentence: \textbackslash n
                          Sentence: [S]. '[L]' in biomedicine is a }}\\
         & \\
         \cline{2-2}
         &  \multirow{2}{*}{\parbox{9cm}{Perform a sentence completion on the following sentence: \textbackslash n
                         Sentence: [L] in medicine is a}}\\
         & \\
         \cline{2-2}
         &  \multirow{2}{*}{\parbox{9cm}{Perform a sentence completion on the following sentence: \textbackslash n
                         Sentence: [L] in biomedicine is a}}\\
         & \\
         \cline{2-2}
         &  \multirow{2}{*}{\parbox{9cm}{Perform a sentence completion on the following sentence: \textbackslash n
                         Sentence: '[L]' in medicine is a }}\\
         & \\
         \cline{2-2}
         &  \multirow{2}{*}{\parbox{9cm}{Perform a sentence completion on the following sentence: \textbackslash n
                         Sentence: '[L]' in biomedicine is a}}\\
         & \\
        \hline
        \multirow{16}{*}{\textit{LLaMA}} 
         &  \multirow{2}{*}{\parbox{9cm}{Perform a sentence completion on the following sentence: 
                          Sentence: [S]. [L] in medicine is a \_\_\_. \textbackslash n The answer is}}\\
         & \\
         \cline{2-2}
         &  \multirow{2}{*}{\parbox{9cm}{Perform a sentence completion on the following sentence: 
                         [S]. [L] in biomedicine is a \_\_\_. \textbackslash n The answer is}}\\
         & \\
         \cline{2-2}
         &  \multirow{2}{*}{\parbox{9cm}{Perform a sentence completion on the following sentence: 
                         [S]. '[L]' in medicine is a \_\_\_. \textbackslash n The answer is }}\\
         & \\
         \cline{2-2}
         &  \multirow{2}{*}{\parbox{9cm}{Perform a sentence completion on the following sentence: 
                          [S]. '[L]' in biomedicine is a \_\_\_. \textbackslash n The answer is }}\\
         & \\
         \cline{2-2}
         &  \multirow{2}{*}{\parbox{9cm}{Perform a sentence completion on the following sentence:
                         [L] in medicine is a \_\_\_. \textbackslash n The answer is}}\\
         & \\
         \cline{2-2}
         &  \multirow{2}{*}{\parbox{9cm}{Perform a sentence completion on the following sentence:
                         [L] in biomedicine is a \_\_\_. \textbackslash n The answer is}}\\
         & \\
         \cline{2-2}
         &  \multirow{2}{*}{\parbox{9cm}{Perform a sentence completion on the following sentence:
                         '[L]' in medicine is a \_\_\_. \textbackslash n The answer is }}\\
         & \\
         \cline{2-2}
         &  \multirow{2}{*}{\parbox{9cm}{Perform a sentence completion on the following sentence: 
                         '[L]' in biomedicine is a \_\_\_. \textbackslash n The answer is}}\\
         & \\
        \hline
    \end{tabular}
\end{table}

\subsection{Task B}
Templates for GeoNames, UMLS, and Schema.Org in zero-shot testing for Task B is for LLMs are presented in \autoref{task-b-templates}.

\begin{table}[]
    \centering
    \caption{The GeoNames, UMLS, and Schema.Org zero-shot testing prompt templates for task B. In the type pairs $(a,b)$ or $(b,a)$, where $a$ is parent and $b$ is child. }
    \label{task-b-templates}
    \begin{tabular}{l   l}
        \hline
        \textbf{LLMs} & \textbf{Prompt Templates} \\
        \hline
         \multirow{8}{*}{\textit{BERT}} 
         & [a] is the superclass of [b]. This statement is a $[MASK]$ . \\
         & [b] is a subclass of [a]. This statement is a $[MASK]$ . \\
         & [a] is the parent class of [b]. This statement is a $[MASK]$ . \\
         & [b] is a child class of [a]. This statement is a $[MASK]$. \\
         & [a] is a supertype of [b]. This statement is a $[MASK]$ . \\
         & [b] is a subtype of [a]. This statement is a $[MASK]$ . \\
         & [a] is an ancestor class of [b]. This statement is a $[MASK]$ . \\
         & [b] is a descendant class of [a]. This statement is a $[MASK]$ . \\
        \hline
        \multirow{8}{*}{\textit{BART}} 
         & [a] is the superclass of [b]. This statement is a $<mask>$ . \\
         & [b] is a subclass of [a]. This statement is a $<mask>$ . \\
         & [a] is the parent class of [b]. This statement is a $<mask>$ . \\
         & [b] is a child class of [a]. This statement is a $<mask>$. \\
         & [a] is a supertype of [b]. This statement is a $<mask>$ . \\
         & [b] is a subtype of [a]. This statement is a $<mask>$ . \\
         & [a] is an ancestor class of [b]. This statement is a $<mask>$ . \\
         & [b] is a descendant class of [a]. This statement is a $<mask>$ . \\
        \hline
         \multirow{8}{*}{\textit{Flan-T5}} 
         & [a] is the superclass of [b]. This statement is a  \\
         & [b] is a subclass of [a]. This statement is a   \\
         & [a] is the parent class of [b]. This statement is a  \\
         & [b] is a child class of [a]. This statement is a  \\
         & [a] is a supertype of [b]. This statement is a \\
         & [b] is a subtype of [a]. This statement is a  \\
         & [a] is an ancestor class of [b]. This statement is a  \\
         & [b] is a descendant class of [a]. This statement is a  \\
        \hline
        \multirow{18}{*}{\textit{BLOOM/GPT-3}}
         &  \multirow{2}{*}{\parbox{9cm}{Identify whether the following statement is true or false: \textbackslash n
                                         Statement: [a] is the superclass of [b]. \textbackslash n This statement is a}}\\
         & \\
         \cline{2-2}
         &  \multirow{2}{*}{\parbox{9cm}{Identify whether the following statement is true or false: \textbackslash n
                                         Statement: [b] is a subclass of [a]. \textbackslash n This statement is a}}\\
         & \\
         \cline{2-2}
         &  \multirow{3}{*}{\parbox{9cm}{Identify whether the following statement is true or false: \textbackslash n
                                         Statement: [a] is the parent class of [b].  \textbackslash n This statement is a }}\\
         & \\
         & \\
         \cline{2-2}
         &  \multirow{2}{*}{\parbox{9cm}{Identify whether the following statement is true or false: \textbackslash n
                                         Statement:  [b] is a child class of [a]. \textbackslash n This statement is a }}\\
         & \\
         \cline{2-2}
          &  \multirow{2}{*}{\parbox{9cm}{Identify whether the following statement is true or false: \textbackslash n
                                         Statement:  [a] is a supertype of [b]. \textbackslash n This statement is a }}\\
         & \\
         \cline{2-2}
         &  \multirow{2}{*}{\parbox{9cm}{Identify whether the following statement is true or false: \textbackslash n
                                         Statement:  [b] is a subtype of [a]. \textbackslash nThis statement is a }}\\
         & \\
         \cline{2-2}
          &  \multirow{2}{*}{\parbox{9cm}{Identify whether the following statement is true or false: \textbackslash n
                                         Statement:  [a] is an ancestor class of [b]. \textbackslash n This statement is a }}\\
         & \\
         \cline{2-2}
          &  \multirow{3}{*}{\parbox{9cm}{Identify whether the following statement is true or false: \textbackslash n
                                         Statement:  [b] is a descendant class of [a]. \textbackslash n This statement is a }}\\
         & \\
         & \\
        \hline
    \end{tabular}
\end{table}

\section{Flan-T5 Training Setups and Hyperparameters}
We finetune Flan-T5 LM on three tasks and evaluate them on all three tasks using zero-shot testing. It involved employing different sources, i.e. WordNet (task A), GeoNames (task A and B), UMLS (the NCI source representing medical sources in task A, B, and C), and schema.org (task B). Considering task A as an 8-shot instance for training we combined samples task B and C training with the condition that only samples that are in the task A 8-shot instances are considered for inclusion. Next, using task-specific prompt templates, Flan-T5 inputs are generated for finetuning. Following, using designed prompt templates Flan-T5 is fine-tuned.  \\

\noindent We utilized a consistent training strategy for all datasets and models, except for a few hyperparameters: batch size and finetuning steps.  All the models were trained using AdamW optimizer with a learning rate of 1e-5. For the Flan-T5-Large model, a batch size of 8 is used during training, while for the Flan-T5-XL model, a batch size of 4 is employed on‌ all datasets. The WordNet and Schema.Org datasets were finetuned for 5 training epochs on both models, similarly, UMLS was finetuned using 10 epochs, while GeoNames was finetuned on Flan-T5-Large for 10 epochs and on Flan-T5-XL for 6 epochs. 

\section{Detailed Results}
The \autoref{task-a-detailed-results} and \autoref{task-a-detailed-results-v2} represent prompt template results across all the templates and LLMs for term typing. While the \autoref{task-b-detailed-results} represents prompt template results across all the templates and LLMs for taxonomy discovery. 

% The detailed results of zero-shot testing and finetuning across seven LLMs reported for ontology learning Task A, term typing in MAP@1.  The results are in percentages. The $*$ denotes finetuning model results.
\begin{table}
\centering
\caption{The detailed results of zero-shot testing and finetuning across seven LLMs reported for ontology learning Task A, term typing in MAP@1.  The results are in percentages. The $*$ denotes finetuning model results. Results for WordNet and GeoName datasets.}\label{task-a-detailed-results}

\begin{tabular}{l l P{0.85cm}P{0.85cm}P{0.85cm}P{0.85cm}P{0.85cm}P{0.85cm}P{0.85cm}P{0.85cm}}
\hline
\multirow{2}{*}{\textbf{Dataset}} & \multirow{2}{*}{\textbf{LLMs}} & \multicolumn{8}{c}{\textbf{Prompt Templates}}  \\
\cline{3-10}
& & \textbf{$t_1$} & \textbf{$t_2$} & \textbf{$t_3$} & \textbf{$t_4$} & \textbf{$t_5$} & \textbf{$t_6$} & \textbf{$t_7$} &  \textbf{$t_8$} \\ \hline
\hline
\multirow{9}{*}{WordNet} 
% & BERT-Large & 2.19 & 9.36 & 9.18 & 19.41 & 4.72 & 19.34 & 9.93 & \textbf{27.85}\\
% & BART-Large & 0.01 & 0.28 & 0.22 & 2.16 & 0.01 & 0.03 & 0.0 & 0.19\\
% & Flan-T5-Large & 0.17 & 19.70 & 5.54 & \textbf{31.26} & 0.0 & 3.03 & 5.70 & 26.80\\
% & Flan-T5-XL & 2.81 & \textbf{40.26} & 17.83 & \textbf{52.21} & 0.01 & 7.75 & 18.47 & 18.85\\
% & BLOOM-1b7 & 54.30 & 61.91 & 64.60 & \textbf{76.00} & 35.99 & 39.64 & 52.06 & 37.82\\ 
% & BLOOM-3b & 29.48 & 48.34 & 62.51 & \textbf{68.35} & 20.68 & 41.04 & 59.11 & 67.68\\ 
% & GPT-3 & 15.32 & 26.55 & \textbf{37.86} & 27.57 & 8.47 & 27.13 & 27.51 & 24.65\\
% & \textit{Flan-T5-Large$^*$} & 73.32 & 76.74 &  54.57 &  \textbf{76.90} &  10.83 &  61.36 &  54.29 &  69.32\\ 
% & \textit{Flan-T5-XL$^*$} & 84.51 &  84.77 &  77.27 &  \textbf{86.28} &  50.23 &  76.46 &  72.38 &  80.51\\ 
&   BERT-Large   &   2.19  &  9.36  &  9.18  &  19.41  &  4.72  &  19.34  &  9.93  &  27.85 \\
% &   PubMedBERT   &   -  &  -  &  -  &  -  &  -  &  -  &  -  &  - \\
&   BART-Large   &   0.01  &  0.28  &  0.22  &  2.16  &  0.01  &  0.03  &  0.0  &  0.19 \\
&   Flan-T5-Large   &   0.17  &  19.70  &  5.54  &  31.26  &  0.0  &  3.03  &  5.70  &  26.80 \\
&   BLOOM-1b7   &   66.83  &  71.53  &  79.20  &  76.84  &  40.08  &  61.96  &  68.39  &  70.03 \\
&   Flan-T5-XL   &   2.81  &  40.26  &  17.83  &  52.21  &  0.01  &  7.75  &  18.47  &  18.85 \\
&   BLOOM-3b   &   63.33  &  75.29  &  79.08  &  77.06  &  37.40  &  65.32  &  68.99  &  71.62 \\
&   LLaMA-7B   &   37.26  &  74.61  &  70.16  &  75.97  &  24.28  &  76.61  &  67.41  &  81.38 \\
&   GPT-3   &   15.32  &  26.55  &  37.86  &  27.57  &  8.47  &  27.13  &  27.51  &  24.65 \\
&   GPT-3.5   &   24.27  &  80.81  &  89.46  &  91.72  &  0.81  &  60.76  &  49.38  &  82.41 \\
&   GPT-4   &   -  &  -  &  -  &  90.11  &  -  &  -  &  -  &  - \\
&  \textit{Flan-T5-Large $^*$}  &   73.32  &  76.74  &  54.57  &  76.90  &  10.83  &  61.36  &  54.29  &  69.32 \\
&  \textit{Flan-T5-XL $^*$}  &   84.51  &  84.77  &  77.27  &  86.28  &  50.23  &  76.46  &  72.38  &  80.51 \\

\hline
\multirow{9}{*}{GeoNames} 
% BERT-Large & \textbf{38.34} &  29.79 &  30.86 &  35.32 &  23.61 &  25.66 &  11.32 &  30.44\\ 
% & BART-Large & 8.47 &  0.57 &  2.23 &  0.98 &  21.48 &  20.51 &  7.83 &  \textbf{23.21}\\ 
% & Flan-T5-Large & 11.55 &  3.57 &  \textbf{13.16} &  4.68 &  9.45 &  6.05 &  8.17 &  7.38\\ 
% & Flan-T5-XL & \textbf{33.81} &  15.71 &  19.77 &  20.78 &  15.36 &  12.41 &  18.43 &  15.82\\
% & BLOOM-1b7 & 7.23 &  8.4 &  2.2 &  6.4 &  23.55 &  20.08 &  22.94 &  \textbf{25.60}\\ 
% & BLOOM-3b & 3.8 &  4.17 &  2.47 &  3.4 &  \textbf{16.79} &  6.95 &  16.68 &  10.78\\ 
% & GPT-3 &  22.42 &  8.72   &  -  &  7.50   &  -   &  -  &   -  & - \\
% & \textit{Flan-T5-Large$^*$} & 15.08 &  15.17 &  14.93 &  15.12 &  15.77 &  16.28 &  15.93 &  \textbf{16.91}\\ 
% & \textit{Flan-T5-XL$^*$} & \textbf{18.35} &  18.12 &  18.12 &  17.91 &  17.26 &  17.32 &  17.45 &  17.64\\
&   BERT-Large   &   38.34  &  29.79  &  30.86  &  35.32  &  23.61  &  25.66  &  11.32  &  30.44 \\
% &   PubMedBERT   &   -  &  -  &  -  &  -  &  -  &  -  &  -  &  - \\
&   BART-Large   &   8.47  &  0.57  &  2.23  &  0.98  &  21.48  &  20.51  &  7.83  &  23.21 \\
&   Flan-T5-Large   &   11.55  &  3.57  &  13.16  &  4.68  &  9.45  &  6.05  &  8.17  &  7.38 \\
&   BLOOM-1b7   &   2.71  &  2.54  &  2.89  &  3.20  &  28.51  &  18.38  &  25.86  &  19.80 \\
&   Flan-T5-XL   &   33.81  &  15.71  &  19.77  &  20.78  &  15.36  &  12.41  &  18.43  &  15.82 \\
&   BLOOM-3b   &   3.76  &  4.70  &  2.64  &  3.43  &  28.84  &  18.08  &  25.64  &  20.71 \\
&   LLaMA-7B   &   29.49  &  14.16  &  25.54  &  15.95  &  13.91  &  9.44  &  17.79  &  16.79 \\
&   GPT-3   &   22.42  &  8.72  &  -  &  7.50  &  -  &  -  &  -  &  - \\
&   GPT-3.5   &   35.00  &  -  &  -  &  -  &  -  &  -  &  -  &  - \\
&   GPT-4   &   43.28  &  -  &  -  &  -  &  -  &  -  &  -  &  - \\
&  \textit{Flan-T5-Large $^*$}  &   15.08  &  15.17  &  14.93  &  15.12  &  15.77  &  16.28  &  15.93  &  16.91 \\
&  \textit{Flan-T5-XL $^*$}  &   18.35  &  18.12  &  18.12  &  17.91  &  17.26  &  17.32  &  17.45  &  17.64 \\
\hline
% \multirow{2}{*}{1} & 0 & 6 & 230 & 35 & 40 & 55 & 25 & 40 & 35 & \\
% & 1 & 5 & 195 & 25 & 50 & 35 & 40 & 45 &  &  \\ \hline
\end{tabular}
\end{table}

\begin{table}
\centering
\caption{The detailed results of zero-shot testing and finetuning across seven LLMs reported for ontology learning Task A, term typing in MAP@1.  The results are in percentages. The $*$ denotes finetuning model results. Results for NCI, SNOMEDCT\_US, and MEDCIN datasets.}\label{task-a-detailed-results-v2}

\begin{tabular}{l l P{0.85cm}P{0.85cm}P{0.85cm}P{0.85cm}P{0.85cm}P{0.85cm}P{0.85cm}P{0.85cm}}
\hline
\multirow{2}{*}{\textbf{Dataset}} & \multirow{2}{*}{\textbf{LLMs}} & \multicolumn{8}{c}{\textbf{Prompt Templates}}  \\
\cline{3-10}
& & \textbf{$t_1$} & \textbf{$t_2$} & \textbf{$t_3$} & \textbf{$t_4$} & \textbf{$t_5$} & \textbf{$t_6$} & \textbf{$t_7$} &  \textbf{$t_8$} \\ \hline
\hline
\multirow{9}{*}{NCI} 
% BERT-Large &      9.94 & 9.76 & 2.61 & 2.90 & \textbf{11.09} & 10.96 & 1.12 & 1.36\\
% & BART-Large &    7.09 & 7.87 & 5.14 & 6.32 & 9.10 & \textbf{9.94} & 7.24 & 8.26\\
% & Flan-T5-Large & 4.59 & 5.06 & 7.53 & \textbf{8.96} & 3.06 & 4.25 & 5.48 & 5.84\\
% & Flan-T5-XL &    4.44 & 5.65 & 7.41 & \textbf{9.83} & 2.12 & 3.29 & 3.87 & 6.28\\
% & BLOOM-1b7 & 7.26 & 7.54 & 8.0 & 8.20 & 10.98 & 11.16 & 12.04 & \textbf{12.37}\\ 
% & BLOOM-3b & 0.19 & 0.11 & 0.54 & \textbf{0.79} & 0.10 & 0.04 & 0.16 & 0.15\\ 
% & GPT-3 & 9.30 & 9.17 & 11.03 & \textbf{12.74} & 9.37 & 8.75 & 9.14 & 9.11\\
% & \textit{Flan-T5-Large$^*$} &30.6 &  31.59 &  31.32 &  \textbf{31.92} &  29.11 &  29.28 &  31.29 &  30.79\\ 
% & \textit{Flan-T5-XL$^*$} & 31.51 &  30.99 &  \textbf{32.78} &  32.05 &  30.01 &  29.7 &  31.76 &  31.35\\ 
&   BERT-Large   &   9.94  &  9.76  &  2.61  &  2.90  &  11.09  &  10.96  &  1.12  &  1.36 \\
  &   PubMedBERT   &   5.87  &  5.36  &  4.52  &  2.79  &  3.36  &  1.61  &  1.33  &  0.65 \\
  &   BART-Large   &   7.09  &  7.87  &  5.14  &  6.32  &  9.10  &  9.94  &  7.24  &  8.26 \\
  &   Flan-T5-Large   &   4.59  &  5.06  &  7.53  &  8.96  &  3.06  &  4.25  &  5.48  &  5.84 \\
  &   BLOOM-1b7   &   12.03  &  12.10  &  11.22  &  12.43  &  10.95  &  10.45  &  11.13  &  11.49 \\
  &   Flan-T5-XL   &   4.44  &  5.65  &  7.41  &  9.83  &  2.12  &  3.29  &  3.87  &  6.28 \\
  &   BLOOM-3b   &   13.77  &  14.35  &  12.94  &  14.41  &  14.26  &  14.06  &  14.92  &  15.56 \\
  &   LLaMA-7B   &   3.78  &  4.05  &  3.24  &  4.77  &  3.67  &  3.92  &  5.25  &  7.71 \\
  &   GPT-3   &   9.30  &  9.17  &  11.03  &  12.74  &  9.37  &  8.75  &  9.14  &  9.11 \\
  &   GPT-3.5   &   11.04  &  9.52  &  14.70  &  14.22  &  8.56  &  8.13  &  12.68  &  11.24 \\
  &   GPT-4   &   -  &  -  &  16.05  &  -  &  -  &  -  &  -  &  - \\
  &  \textit{Flan-T5-Large $^*$}  &   30.60  &  31.59  &  31.32  &  31.92  &  29.11  &  29.28  &  31.29  &  30.79 \\
  &  \textit{Flan-T5-XL $^*$}  &   31.51  &  30.99  &  32.78  &  32.05  &  30.01  &  29.70  &  31.76  &  31.35 \\
\hline
\multirow{9}{*}{SNOMEDCT} 
% BERT-Large      & 19.83 & 8.02  & 1.06  & 0.12 & \textbf{21.10} & 12.76 & 0.45 & 0.04\\
% & BART-Large    & 19.16 & \textbf{19.81} & 4.16  & 4.04 & 17.54 & 17.89 & 10.06 & 9.43\\
% & Flan-T5-Large & 19.26 & 19.89 & 21.04 & \textbf{24.32} & 8.07 & 8.90 & 11.54 & 12.92\\
% & Flan-T5-XL    & 25.21 & 26.23 & 30.09 & \textbf{31.65} & 7.21 & 8.22 & 15.58 & 17.22\\
% & BLOOM-1b7 & 17.65 & 14.92 & 10.95 & 9.73 & 28.87 & 27.21 & \textbf{32.32} & 30.82\\ 
% & BLOOM-3b & 0.99 & 0.47 & \textbf{2.16} & 2.09 & 1.39 & 0.66 & 0.91 & 0.52\\ 
% & GPT-3 & 21.06 & 20.33 & 22.73 & \textbf{24.36} & 19.2 & 18.99 & 20.20 & 20.09\\
% & \textit{Flan-T5-Large$^*$}  & 32.27 &  31.99 &  31.56 &  31.36 &  32.0 &  31.5 &  \textbf{33.39} &  33.05\\
% & \textit{Flan-T5-XL$^*$} & \textbf{43.39} &  42.03 &  42.76 &  41.75 &  40.89 &  40.31 &  42.6 &  42.48\\ 
&   BERT-Large   &   19.83  &  8.02  &  1.06  &  0.12  &  21.10  &  12.76  &  0.45  &  0.04 \\
  &   PubMedBERT   &   28.48  &  22.47  &  13.91  &  5.70  &  7.96  &  3.58  &  2.29  &  1.51 \\
  &   BART-Large   &   19.16  &  19.81  &  4.16  &  4.04  &  17.54  &  17.89  &  10.06  &  9.43 \\
  &   Flan-T5-Large   &   19.26  &  19.89  &  21.04  &  24.32  &  8.07  &  8.90  &  11.54  &  12.92 \\
  &   BLOOM-1b7   &   32.43  &  37.02  &  13.78  &  19.97  &  29.48  &  30.40  &  31.24  &  33.86 \\
  &   Flan-T5-XL   &   25.21  &  26.23  &  30.09  &  31.65  &  7.21  &  8.22  &  15.58  &  17.22 \\
  &   BLOOM-3b   &   34.26  &  37.69  &  27.18  &  27.87  &  31.06  &  32.21  &  33.29  &  35.47 \\
  &   LLaMA-7B   &   7.56  &  6.75  &  7.89  &  8.06  &  10.74  &  10.80  &  13.15  &  13.81 \\
  &   GPT-3   &   21.06  &  20.33  &  22.73  &  24.36  &  19.20  &  18.99  &  20.20  &  20.09 \\
  &   GPT-3.5   &   21.81  &  17.99  &  25.02  &  24.50  &  18.24  &  15.71  &  22.71  &  19.87 \\
  &   GPT-4   &   -  &  -  &  22.36  &  27.83  &  -  &  -  &  -  &  - \\
  &  \textit{Flan-T5-Large $^*$}  &   32.27  &  31.99  &  31.56  &  31.36  &  32.00  &  31.50  &  33.39  &  33.05 \\
  &  \textit{Flan-T5-XL $^*$}  &   43.39  &  42.03  &  42.76  &  41.75  &  40.89  &  40.31  &  42.60  &  42.48 \\

\hline
\multirow{9}{*}{MEDCIN} 
% & BERT-Large    & 7.33  & 1.25  & 0.14  & 0.05 & \textbf{8.71} & 1.19 & 0.08 & 0.01\\
% & BART-Large    & 11.67 & \textbf{12.65} & 2.27  & 2.31 & 9.40 & 9.22 & 5.47 & 4.82\\
% & Flan-T5-Large & 9.30  & 8.08  & 10.97 & \textbf{12.96} & 2.89 & 3.59 & 6.71 & 6.78\\
% & Flan-T5-XL    & 15.24 & 15.89 & 18.04 & \textbf{18.51} & 4.47 & 5.44 & 11.04 & 11.09\\
% & BLOOM-1b7 & 6.94 & 4.61 & 1.60 & 1.37 & 12.05 & 10.59 & \textbf{16.87} & 14.25\\ 
% & BLOOM-3b & 0.04 & 0.02 & \textbf{0.11} & 0.06 & 0.03 & 0.02 & 0.05 & 0.04\\ 
% & GPT-3 & 22.4 & 22.56 & \textbf{25.72} & 24.91 & 19.75 & 17.8 & 19.92 & 18.57\\
% & \textit{Flan-T5-Large$^*$} & \textbf{38.37} &  36.37 &  37.43 &  35.86 &  31.26 &  30.0 &  33.11 &  31.91\\
% & \textit{Flan-T5-XL$^*$} & \textbf{51.80} &  50.90 &  \textbf{51.80} &  51.16 &  47.88 &  45.38 &  49.86 &  49.09\\
&   BERT-Large   &   7.33  &  1.25  &  0.14  &  0.05  &  8.71  &  1.19  &  0.08  &  0.01 \\
  &   PubMedBERT   &   15.62  &  9.71  &  5.20  &  1.58  &  5.68  &  2.32  &  1.27  &  0.61 \\
  &   BART-Large   &   11.67  &  12.65  &  2.27  &  2.31  &  9.40  &  9.22  &  5.47  &  4.82 \\
  &   Flan-T5-Large   &   9.30  &  8.08  &  10.97  &  12.96  &  2.89  &  3.59  &  6.71  &  6.78 \\
  &   BLOOM-1b7   &   27.58  &  28.67  &  2.70  &  4.97  &  26.38  &  28.76  &  26.89  &  26.69 \\
  &   Flan-T5-XL   &   15.24  &  15.89  &  18.04  &  18.51  &  4.47  &  5.44  &  11.14  &  11.09 \\
  &   BLOOM-3b   &   23.05  &  28.31  &  14.39  &  10.82  &  22.58  &  24.23  &  27.30  &  29.81 \\
  &   LLaMA-7B   &   3.40  &  2.80  &  3.37  &  3.73  &  4.90  &  4.47  &  3.17  &  3.80 \\
  &   GPT-3   &   22.40  &  22.56  &  25.72  &  24.91  &  19.75  &  17.80  &  19.92  &  18.57 \\
  &   GPT-3.5   &   22.51  &  22.06  &  23.92  &  23.58  &  20.46  &  19.84  &  22.37  &  20.23 \\
  &   GPT-4   &   -  &  -  &  21.25  &  23.61  &  -  &  -  &  -  &  - \\
  &  \textit{Flan-T5-Large $^*$}  &   38.37  &  36.37  &  37.43  &  35.86  &  31.26  &  30.00  &  33.11  &  31.91 \\
  &  \textit{Flan-T5-XL $^*$}  &   51.80  &  50.90  &  51.80  &  51.16  &  47.88  &  45.38  &  49.86  &  49.09 \\
\hline
% \multirow{2}{*}{1} & 0 & 6 & 230 & 35 & 40 & 55 & 25 & 40 & 35 & \\
% & 1 & 5 & 195 & 25 & 50 & 35 & 40 & 45 &  &  \\ \hline
\end{tabular}
\end{table}

\begin{table}
\centering
\caption{The detailed results of zero-shot testing and finetuning across seven LLMs reported for ontology learning Task B, type taxonomy discovery in F1-score.  The results are in percentages. The $*$ denotes finetuning model results.}\label{task-b-detailed-results}

\begin{tabular}{l l P{0.85cm}P{0.85cm}P{0.85cm}P{0.85cm}P{0.85cm}P{0.85cm}P{0.85cm}P{0.85cm}}
\hline
\multirow{2}{*}{\textbf{Dataset}} & \multirow{2}{*}{\textbf{LLMs}} & \multicolumn{8}{c}{\textbf{Prompt Templates}}  \\
\cline{3-10}
& & \textbf{$t_1$} & \textbf{$t_2$} & \textbf{$t_3$} & \textbf{$t_4$} & \textbf{$t_5$} & \textbf{$t_6$} & \textbf{$t_7$} &  \textbf{$t_8$} \\ \hline
\multirow{9}{*}{GeoNames} 
% & BERT-Large & 41.0 &  51.69 &  40.55 &  48.7 &  37.16 &  41.07 &  41.7 &  \textbf{54.54}\\ 
% & BART-Large & 38.11 &  41.03 &  40.55 &  52.5 &  39.09 &  45.8 &  36.67 &  \textbf{55.4}\\ 
% & Flan-T5-Large & \textbf{59.63} &  48.24 &  54.08 &  48.24 &  44.4 &  51.3 &  36.4 &  38.44\\ 
% & Flan-T5-XL & 49.37 &  44.05 &  45.09 &  \textbf{52.41} &  43.92 &  46.34 &  49.98 &  44.29\\ 
% & BLOOM-1b7 & 33.16 &  31.04 &  33.16 &  32.83 &  33.77 &  33.53 &  \textbf{36.67} &  32.92\\ 
% & BLOOM-3b & 35.85 &  39.12 &  53.92 &  30.22 &  35.62 &  33.6 &  \textbf{48.26} &  37.73\\ 
% & GPT-3 & 43.43 &  51.74 &  42.7 &  \textbf{53.2} &  46.04 &  52.56 &  45.49 &  52.62\\ 
% & \textit{Flan-T5-Large$^*$} & 42.53 &  59.4 &  40.29 &  \textbf{62.46} &  46.03 &  57.41 &  42.49 &  62.04\\ 
% & \textit{Flan-T5-XL$^*$} & 48.41 &  34.8 &  55.23 &  46.96 &  57.48 &  36.29 &  \textbf{59.05} &  49.26\\ 
&   BERT-Large   &   41.00  &  51.69  &  40.55  &  48.70  &  37.16  &  41.07  &  41.70  &  54.54 \\
  % &   PubMedBERT   &   -  &  -  &  -  &  -  &  -  &  -  &  -  &  - \\
  &   BART-Large   &   38.11  &  41.03  &  40.55  &  52.50  &  39.09  &  45.80  &  36.67  &  55.40 \\
  &   Flan-T5-Large   &   59.63  &  48.24  &  54.08  &  48.24  &  44.40  &  51.30  &  36.40  &  38.44 \\
  &   BLOOM-1b7   &   33.16  &  31.04  &  33.16  &  32.83  &  33.77  &  33.53  &  36.67  &  32.92 \\
  &   Flan-T5-XL   &   49.37  &  44.05  &  45.09  &  52.41  &  43.92  &  46.34  &  49.98  &  44.29 \\
  &   BLOOM-3b   &   35.85  &  39.12  &  53.92  &  30.22  &  35.62  &  33.60  &  48.26  &  37.73 \\
  &   LLaMA-7B   &   33.49  &  33.49  &  33.49  &  33.49  &  33.49  &  33.49  &  33.49  &  33.49 \\
  &   GPT-3   &   43.43  &  51.74  &  42.70  &  53.20  &  46.04  &  52.56  &  45.49  &  52.62 \\
  &   GPT-3.5   &   59.40  &  47.79  &  67.78  &  41.95  &  48.02  &  51.72  &  45.25  &  43.86 \\
  &   GPT-4   &   38.56  &  52.46  &  34.00  &  38.89  &  44.06  &  55.43  &  33.78  &  36.23 \\
  & \textit{Flan-T5-Large$^*$}   &   42.53  &  59.40  &  40.29  &  62.46  &  46.03  &  57.41  &  42.49  &  62.04 \\
  & \textit{Flan-T5-XL$^*$}  &   48.41  &  34.80  &  55.23  &  46.96  &  57.48  &  36.29  &  59.05  &  49.26 \\
\hline
\multirow{9}{*}{UMLS} 
% & BERT-Large & \textbf{48.21} &  38.84 &  41.46 &  40.41 &  45.88 &  40.91 &  41.04 &  42.92\\ 
% & BART-Large & 36.02 &  48.21 &  41.42 &  \textbf{49.9} &  39.37 &  47.47 &  42.39 &  45.46\\ 
% & Flan-T5-Large & 47.55 &  51.22 &  \textbf{55.32} &  40.94 &  49.45 &  50.87 &  44.23 &  42.9\\ 
% & Flan-T5-XL & \textbf{64.25} &  46.53 &  51.0 &  41.54 &  60.07 &  42.83 &  51.25 &  41.18\\ 
% & BLOOM-1b7 & 33.71 &  36.18 &  33.71 &  \textbf{38.26} &  33.71 &  35.89 &  33.27 &  33.6\\ 
% & BLOOM-3b & 33.16 &  37.23 &  34.82 &  35.77 &  33.16 &  35.89 &  33.05 &  \textbf{37.48}\\ 
% & GPT-3 & \textbf{51.58} &  49.41 &  49.86 &  42.9 &  50.57 &  46.07 &  45.36 &  46.72\\ 
% & \textit{Flan-T5-Large$^*$} & 37.17 &  48.66 &  36.07 &  42.12 &  48.39 &  46.65 &  \textbf{53.42} &  35.97\\ 
% & \textit{Flan-T5-XL$^*$} & 63.69 &  50.04 &  36.91 &  41.34 &  78.12 &  50.12 &  \textbf{79.25} &  39.27\\
  &   BERT-Large   &   48.21  &  38.84  &  41.46  &  40.41  &  45.88  &  40.91  &  41.04  &  42.92 \\
  &   PubMedBERT   &   33.71  &  33.71  &  33.71  &  33.71  &  33.71  &  33.71  &  33.71  &  33.71 \\
  &   BART-Large   &   36.02  &  48.21  &  41.42  &  49.90  &  39.37  &  47.47  &  42.39  &  45.46 \\
  &   Flan-T5-Large   &   47.55  &  51.22  &  55.32  &  40.94  &  49.45  &  50.87  &  44.23  &  42.90 \\
  &   BLOOM-1b7   &   33.71  &  36.18  &  33.71  &  38.26  &  33.71  &  35.89  &  33.27  &  33.60 \\
  &   Flan-T5-XL   &   64.25  &  46.53  &  51.00  &  41.54  &  60.07  &  42.83  &  51.25  &  41.18 \\
  &   BLOOM-3b   &   33.16  &  37.23  &  34.82  &  35.77  &  33.16  &  35.89  &  33.05  &  37.48 \\
  &   LLaMA-7B   &   32.94  &  32.94  &  32.94  &  32.94  &  32.94  &  32.94  &  32.94  &  32.94 \\
  &   GPT-3   &   51.58  &  49.41  &  49.86  &  42.90  &  50.57  &  46.07  &  45.36  &  46.72 \\
  &   GPT-3.5   &   61.38  &  70.38  &  63.91  &  66.82  &  63.14  &  67.27  &  56.64  &  64.41 \\
  &   GPT-4   &   41.19  &  76.99  &  42.55  &  63.88  &  50.28  &  78.11  &  36.59  &  60.72 \\
  &    \textit{Flan-T5-Large$^*$}    &   37.17  &  48.66  &  36.07  &  42.12  &  48.39  &  46.65  &  53.42  &  35.97 \\
  &  \textit{Flan-T5-XL$^*$}  &   63.69  &  50.04  &  36.91  &  41.34  &  78.12  &  50.12  &  79.25  &  39.27 \\
  
\hline
\multirow{9}{*}{schema.org}  
% & BERT-Large & 43.85 &  41.17 &  \textbf{44.06} &  43.2 &  43.7 &  40.05 &  42.15 &  43.72\\ 
% & BART-Large & 34.62 &  38.69 &  39.28 &  \textbf{52.9} &  38.2 &  41.17 &  43.26 &  42.74\\ 
% & Flan-T5-Large & 46.98 &  49.92 &  46.11 &  \textbf{54.78} &  40.27 &  54.47 &  42.06 &  47.93\\ 
% & Flan-T5-XL & \textbf{42.70} &  33.45 &  33.59 &  42.76 &  36.69 &  34.04 &  33.75 &  36.45\\ 
% & BLOOM-1b7 & 33.39 &  47.83 &  33.39 &  39.77 &  38.92 &  \textbf{48.56} &  44.35 &  39.57\\ 
% & BLOOM-3b & 41.64 &  47.16 &  47.98 &  45.25 &  39.73 &  40.75 &  \textbf{51.28} &  48.73\\ 
% & GPT-3 & 49.64 &  49.28 &  \textbf{50.97} &  48.03 &  47.19 &  48.63 &  48.87 &  49.48\\ 
% &  \textit{Flan-T5-Large$^*$} & 35.35 &  85.43 &  29.82 &  89.24 &  41.3 &  \textbf{91.68} &  42.46 &  56.39\\ 
% &  \textit{Flan-T5-XL$^*$} & 91.06 &  57.46 &  74.68 &  65.32 &  91.54 &  50.63 &  \textbf{91.70} &  33.33\\ 
&   BERT-Large   &   43.85  &  41.17  &  44.06  &  43.20  &  43.70  &  40.05  &  42.15  &  43.72 \\
% &   PubMedBERT   &   -  &  -  &  -  &  -  &  -  &  -  &  -  &  - \\
&   BART-Large   &   34.62  &  38.69  &  39.28  &  52.90  &  38.20  &  41.17  &  43.26  &  42.74 \\
&   Flan-T5-Large   &   46.98  &  49.92  &  46.11  &  54.78  &  40.27  &  54.47  &  42.06  &  47.93 \\
&   BLOOM-1b7   &   33.39  &  47.83  &  33.39  &  39.77  &  38.92  &  48.56  &  44.35  &  39.57 \\
&   Flan-T5-XL   &   42.70  &  33.45  &  33.59  &  42.76  &  36.69  &  34.04  &  33.75  &  36.45 \\
&   BLOOM-3b   &   41.64  &  47.16  &  47.98  &  45.25  &  39.73  &  40.75  &  51.28  &  48.73 \\
&   LLaMA-7B   &   33.37  &  33.37  &  33.37  &  33.37  &  33.37  &  33.37  &  33.37  &  33.37 \\
&   GPT-3   &   49.64  &  49.28  &  50.97  &  48.03  &  47.19  &  48.63  &  48.87  &  49.48 \\
&   GPT-3.5   &   56.84  &  74.38  &  58.52  &  70.16  &  53.35  &  72.35  &  54.16  &  71.03 \\
&   GPT-4   &   58.47  &  72.82  &  65.83  &  63.30  &  50.56  &  74.24  &  57.45  &  63.69 \\
&   \textit{Flan-T5-Large$^*$}   &   35.35  &  85.43  &  29.82  &  89.24  &  41.30  &  91.68  &  42.46  &  56.39 \\
&   \textit{Flan-T5-XL$^*$}   &   91.06  &  57.46  &  74.68  &  65.32  &  91.54  &  50.63  &  91.70  &  33.33 \\
\hline

\end{tabular}
\end{table}
\end{document}